\documentclass[manuscript,screen]{acmart}
\usepackage{amsmath,amsfonts}
\usepackage{algorithmic}
\usepackage{graphicx}
\usepackage{textcomp}
\usepackage{url}
\usepackage{tabularx}
\usepackage{url}
\usepackage{booktabs}
\usepackage{multirow}
\usepackage{longtable}

\AtBeginDocument{%
  \providecommand\BibTeX{{%
    \normalfont B\kern-0.5em{\scshape i\kern-0.25em b}\kern-0.8em\TeX}}}

\setcopyright{acmcopyright}

\acmJournal{CSUR}
\acmYear{2020} 
\acmVolume{1} 
\acmNumber{1} 
\acmArticle{1} 
\acmMonth{1} 

\begin{document}

\title{Dataset of Fake News Detection and Fact Verification: A Survey}

\author{Taichi Murayama}
\email{murayama.taichi.mk1@is.naist.jp}
\orcid{0000-0003-1148-711X}
\affiliation{%
  \institution{NARA Institute of Science and Technology}
  \streetaddress{Takayamacho 8916-5}
  \city{Ikoma}
  \state{Nara}
  \country{Japan}
  \postcode{630-0192}
}

\renewcommand{\shortauthors}{T. Murayama}

\begin{abstract}
The rapid increase in fake news, which causes significant damage to society, triggers many fake news related studies, including the development of fake news detection and fact verification techniques.
The resources for these studies are mainly available as public datasets taken from Web data.
We surveyed 118 datasets related to fake news research on a large scale from three perspectives: (1) fake news detection, (2) fact verification, and (3) other tasks; for example, the analysis of fake news and satire detection.
We also describe in detail their utilization tasks and their characteristics.
Finally, we highlight the challenges in the fake news dataset construction and some research opportunities that addresses these challenges.
Our survey facilitates fake news research by helping researchers find suitable datasets without reinventing the wheel, and thereby, improves fake news studies in depth.
\end{abstract}



\maketitle
\section{Introduction}
Fake news has caused significant damage to various aspects of society.
For example, in stock markets, a false report about the bankruptcy of the United Airlines’ parent company in 2008 caused the company's stock price to drop by 76\% in a few minutes; it closed at 11\% below the previous day, with a negative effect persisting for more than six days~\cite{united_fake}.
During the 2016 US presidential election, 529 different low-credibility statements~\cite{president1} were spread on Twitter; 
25\% of the news outlets linked from tweets, which were either fake or extremely biased and supporting Trump or Clinton, potentially influencing the election~\cite{president2}.
Fake news may cause significant impact on real events, so much so that the ``Pizzagate'' conspiracy that started on Reddit led to a real shooting~\cite{pizzagate}.
In addition to stock markets and political events, it is the same situation in public health.
Fake news on infectious diseases, such as Ebola~\cite{ebola}, yellow fever~\cite{yellow}, and Zika~\cite{zika}, appears to be spreading on the Internet.
The COVID-19 pandemic from 2020 especially continues to produce news from many doubtful sources~\cite{covid_fakenews, covid_fakenews2}.
The number of English-language fact-check articles increased by more than 900\% during January to March 2020~\cite{covid_factcheck}.
Fake news has had a significant impact on events in various countries: Brexit in Europe~\cite{brexit1,brexit2}, the salt panic in China~\cite{salt}, deadly violence between the two groups, ethnic Oromos and ethnic Somalis, in Ethiopia~\cite{ethiopia}, and natural disasters such as the East Japan Great Earthquake of 2011~\cite{tohoku1, tohoku2}, and the Chile earthquake in 2010~\cite{disaster1}.

While fake news is not a new phenomenon, it has become a significant problem because of social media.
Compared to traditional news media such as newspapers and television, it is now easy to share sensational news with many people using social media such as Facebook, Twitter, and Weibo~\cite{allcott2017social}.
The Pew Research Center reports that about a half of the US adults mainly receive news from social media~\cite{pewrecearch}.
Social media provides an ideal environment for communication and information acquisition, and encourages users to share information without distance being a barrier between individuals.
Meanwhile, it also becomes the mechanism for accelerated fake news dissemination due to the existence of an echo chamber effect~\cite{jamieson2008echo, disaster1, facebook_failed}.
In particular, younger and older people tend to believe in fake news, which in turn, becomes an incentive to create, publish, and spread fake news for potentially substantial political and economic benefits~\cite{forbe2016}.
The prevalence of fake news in social media has the potential to break the trustworthiness of online journalism and cause widespread panic, and poses a major problem in the aforementioned examples.

Serious social concern about fake news prompts researchers to combat fake news using methods such as detection, verification, mitigation, and analysis.
In this regard, there have been attempts to survey and summarize the literature on fake news research.
Many tutorials at international conferences have supported fake news research~\cite{zhou2019fake,zafarani2019fake,nakov2020fact,giachanou2020battle,nakov2021fake}.
In addition, some workshops (for example, FEVER~\cite{fever_workshop}, Fake News Challenge~\cite{fakenewschallenge}, RumorEval~\cite{Rumoreval2019}, CheckThat!~\cite{checkthat2021}, and Constraint~\cite{constraint_workshop}) organize competitions to improve the techniques of fake news detection and fact verification.
These attempts have helped researchers, especially those with no previous experience in the area, to work in this field.
Most of existing literature mainly summarizes and introduces methods and technologies for detection and verification. 
However, they do not focus much on the dataset used for these methods and technologies.

We aim to provide summaries of various datasets related to fake news.
This study comprehensively introduces 118 kinds of datasets related to fake news, significantly more than 27 presented in a previous survey paper~\cite{d2021fake} focusing on the dataset.
The study can facilitate fake news research by helping researchers find the suitable dataset without ``reinventing the wheel,'' and improve fake news studies in depth.
Before we provide a summary of our work in Section 1.3, we describe the definition of ``fake news'' in Section 1.1 and related concepts of fake news in Section 1.2.

\subsection{Definition of fake news}
Claire Wardle, the co-founder and leader of the First Draft~\cite{first_draft}, announced that the term fake news is woefully inadequate to describe the issues related to it, and distinguishes between three types of information content problems: misinformation, disinformation, and malinformation ~\cite{cnn_clair}.
Therefore, it is not easy to construct a general definition of ``fake news'' reflecting the current diverse circumstances.

``Fake news is false news'' is a broad definition of fake news~\cite{zhou2020survey}.
Similarly, Lazer et al.~\cite{lazer2018science} describes ``fake news is fabricated information that mimics news media content in form but not in organizational process or intent.''
This definition is a broad definition that emphasizes only informational authenticity and does not consider informational intention.
It allows us to cover different types of fake news identified by their motive or intent, such as satire and parody~\cite{rubin2015deception}.
There are few studies~\cite{science2018,sharma2019combating,jin2016news} leveraging the definition.

Most research emphasizes ``intention'' in the definition of fake news, as a narrow definition.
Allcott and Shu et al.~\cite{shu2017fake,allcott2017social} define fake news as ``a news article that is \textbf{intentionally} and verifiably false.''
Zhang et al.~\cite{zhang2020overview} describes that ``fake news refers to all kinds of false stories or news that are mainly published and distributed on the Internet, in order to \textbf{purposely} mislead, befool or lure readers for financial, political or other gains.''
Other studies ~\cite{mustafaraj2017fake,conroy2015automatic,potthast2017stylometric} have also emphasized the intention in the definition of fake news.
However, we believe that most of these studies related to fake news do not completely follow this definition.
Researchers use fake news datasets labeled as either fake or not fake based on the judgment of fact-checking sites, most of which do not consider the creator's intention as it is difficult to judge whether each piece of fake news is created with the intention of misleading readers.
We refer to \cite{tandoc2018defining} for a more detailed discussion of the range of meanings.

\subsection{Related concepts of fake news}
Under the common practice of conveying false information, the term ``fake news'' is closely related to several concepts: satire, rumor, clickbait, and so on.
There are salient differences among them in terms of the context of usage and for different propagation purposes.

\begin{itemize}
    \item \textbf{Misinformation} is false information that is inaccurate or misleading in a macro aspect~\cite{lazer2018science,misinfo}.
    It spreads unintentionally because of honest mistakes~\cite{hernon1995disinformation} or knowledge updation without the purpose of misleading.
    
    \item \textbf{Disinformation} is false information that misleads others intentionally for a certain purpose (e.g., to deceive people~\cite{kumar2016disinformation}, to promote a biased agenda~\cite{volkova2017separating}), which is close to a narrow definition of fake news.
    In contrast to misinformation, it spreads because of a deliberate attempt to deceive or mislead others~\cite{hernon1995disinformation}. 
    The survey by Guo et al.~\cite{guo2020future} divides false news into two categories: misinformation and disinformation.
    Sometimes, the term ``deception'' is treated in a similar way to disinformation~\cite{zhou2020survey}.
    
    \item \textbf{Rumor} is an unverified and relevant information being circulated, and it could later be confirmed as true, false, or left unconfirmed~\cite{buntain2017automatically,peterson1951rumor}.
    It could spread from one user to another.
    Before the term ``fake news'' became popular, the task of classifying news as false or not was called ``rumor detection,'' related to which there has been a lot of research.
    Zubiaga et al. summary the studies on rumor~\cite{zubiaga2018detection}.
    
    \item \textbf{Hoaxes} are deliberately fabricated information made to masquerade as the truth~\cite{kumar2016disinformation}.
    It often causes serious material damage to the victim because it includes relatively complex and large-scale fabrications~\cite{rubin2015deception,brunvand2006american}.
    
    \item \textbf{Satire}, which contains a lot of irony and humor, is written with the purpose of entertaining or criticizing the readers~\cite{brummette2018read}.
    These news articles are frequently published at some sites, such as SatireWire.com~\cite{satirewire} and The Onion~\cite{onion}.
    It could be harmful if satire news, ignoring the context, is shared by many people.
    ``Parody'' is a similar concept to satire, but there are differences, where parody uses non-factual information to inject humor~\cite{tandoc2018defining}.

    \item \textbf{Hyperpartisan} news is extremely one-sided or biased news in a political context~\cite{potthast2017stylometric}.
    In itself, biased does not mean being fake; however, some papers~\cite{icsss2,zannettou2018gab} report that it has a high possibility of being false in hyperpartisan news networks and that false information is spread widely in the alt-right community, such as 4chan's /pol/ board~\cite{4ch} and Gab~\cite{gab}.
    
    \item \textbf{Propaganda} is a form of persuasion that attempts to influence the emotions, attitudes, opinions, and actions of specified target audiences for political, ideological, and religious purposes through the controlled transmission of one-sided messages~\cite{jowett2018propaganda,lumezanu2012bias}.
    Recently, it has often been used to influence election results and opinions in a political context.

    
    \item \textbf{Spam} is fabricated information that ranges from self-promotions to false announcements of the product.
    In review sites, spam provides positive reviews to unfairly promote products or unjustified negative reviews to competing products in order to damage their reputation~\cite{jindal2008opinion}.
    In the social ecosystem, spam targets users to disseminate malware and commercial spam messages to promote affiliate websites~\cite{lee2010uncovering}.

    \item \textbf{Clickbait} is a story with an eye-catching headline that is intended to attract the traffic and benefit from advertising revenue~\cite{volkova2017separating,chen2015misleading}.
    Because of the discrepancy between content and headline is the main component in clickbait, it is one of the least severe types of false information.
\end{itemize}

These term lists are based on \cite{terms1,zhou2020survey,guo2020future} and extended to build upon the existing literature.
There are no formal definitions for these terms, similar to the term ``fake news.''
In addition, the term lists should not be treated as an exhaustive representation of the false information ecosystem (e.g., half-truth~\cite{zannettou2019web} and factoid~\cite{wu2020providing}). 
In the following sections, the usage of these terms in this study complies with reference papers.

\subsection{An overview of this survey}
\begin{figure}
\begin{tabular}{c}
  \begin{minipage}{1\textwidth}
    \centering
    \captionsetup{width=12cm}
    \includegraphics[width=10cm]{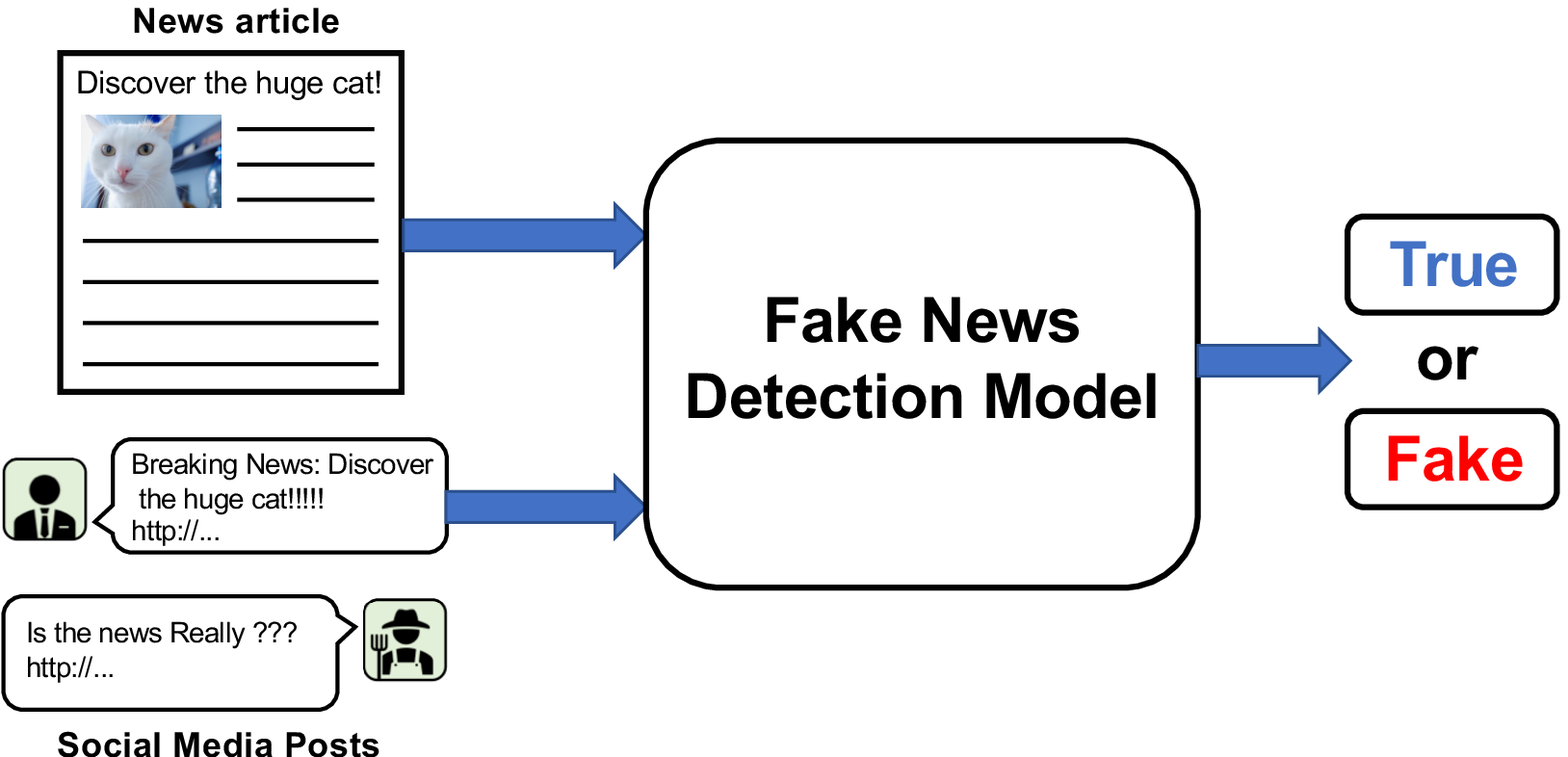}
    \caption{\textbf{Fake news detection task} classifies whether each news item is fake, by getting from, as input data, text information such as the news content or social media posts including the news.}
    \vspace{0.5cm}
    \label{fig1}
  \end{minipage}\\
   \begin{minipage}{1\textwidth}
    \centering
    \captionsetup{width=12cm}
    \includegraphics[width=11cm]{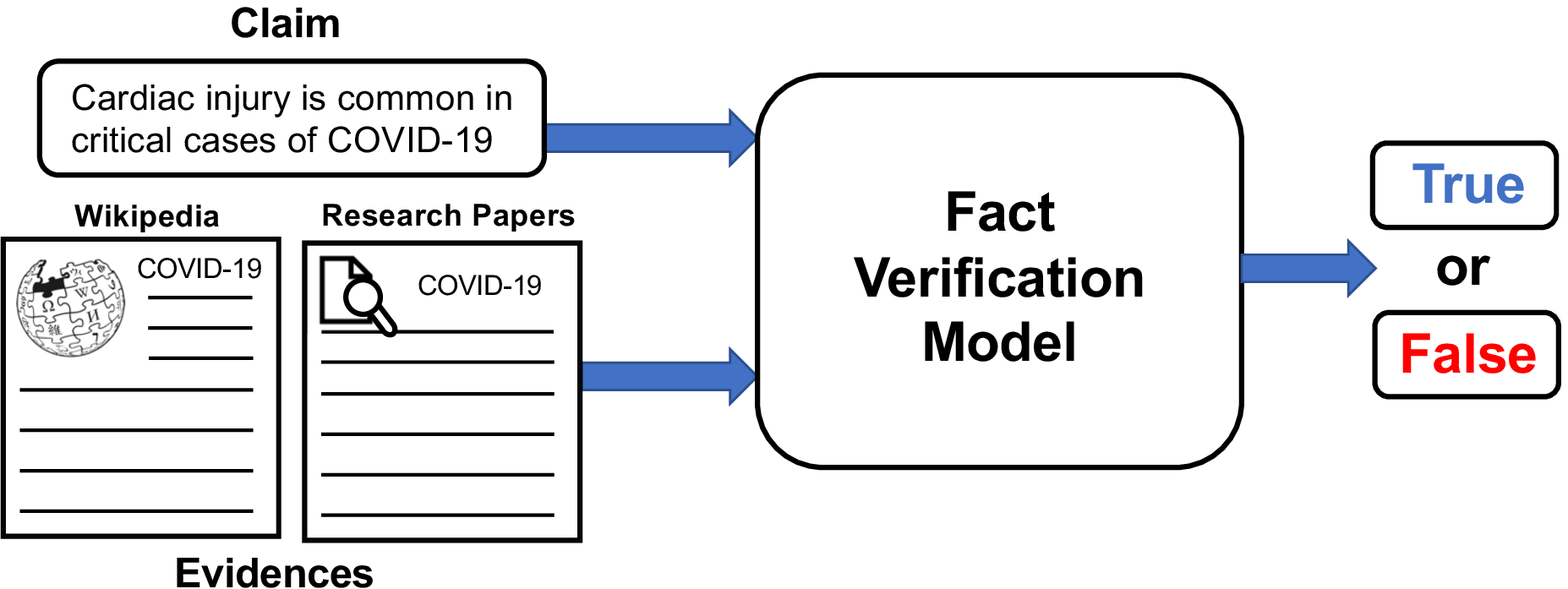}
    \caption{\textbf{Fact verification task} is to make the decision as to whether a claim is correct, based on the explicitly-available evidence, such as Wikipedia articles and research papers.}
    \label{fig2}
  \end{minipage}
\end{tabular}
\end{figure}

Our survey aims to present a comprehensive summary of existing datasets, that are publicly available or easy to obtain, to quantitatively analyze and study fake news.
Methods for fake news detection and fact verification, which many existing surveys comprehensively introduce, are not our target.
We expect the summary to be an index for further research, quickly identifying previous datasets suitable for their demand or gaps between the existing literature and their demand.
We present datasets related to fake news from three aspects that are categorized by research content: (1) fake news detection, classifying fake news from text and other information, (2) fact verification, judging whether a claim is true from evidence, and (3) other datasets related to fake news, which are developed neither for fake news detection nor for fact verification (e.g., fake news analysis, fake media bias analysis, or a novel task for fake news).
An overview of the fake news detection and fact verification task is shown in Figure~\ref{fig1} and Figure~\ref{fig2}.

Our survey mainly focuses on datasets that include text information in fake news (e.g., misinformation, rumor, hoaxes, and satire).
It does not cover specific topics, despite their close relationship to fake news.
For example, datasets that mainly include videos with false information such as Deepfakes~\cite{floridi2018artificial,kopev2019detecting, guera2018deepfake}, utilizing deep learning models to create audio and video of real people saying and doing things they never said or did, are not covered.
Refer to \cite{mirsky2021creation,alam2021survey} for more comprehensive information on detecting multimodal fake news and deepfakes.
In addition, datasets focusing on spam reviews or fake reviews, which are written to promote or demote a few target products or services, are not covered.
Some studies~\cite{heydari2015detection, patel2018survey} provides a comprehensive summary of this topic.

We first review existing survey papers and major fact-checking organizations that are useful in building fake news datasets in Section 2.
We then present summaries of datasets related to fake news from three aspects: fake news detection, fact verification, and other datasets related to fake news in Sections 3--5.
Section 6 discusses some problems in fake news dataset construction that need to be resolved in the future.
We conclude the survey in Section 7.

\section{Related Work}

\subsection{Survey of researches about fake news}
There have been many attempts to survey the literature on fake news research.
Before the term "fake news" became known among researchers, there was a large number of research on rumors and credibility of information~\cite{zubiaga2018detection,li2016survey}.
Shu et al.~\cite{shu2017fake} provide a summary of early fake news research.
They present a comprehensive review of detecting fake news on social media, including fake news characterizations of psychology and social theories and on existing algorithms from a data mining perspective, evaluation metrics, and representative datasets.
Kumar et al.~\cite{kumar2018false} summarize fake news research based on two categories: opinion-based, where a unique ground truth does not exist as in the case of product reviews, and fact-based that consists of lies about entities with unique ground truth values.
Zannettou et al.~\cite{zannettou2019web} report a comprehensive overview of existing research from the perspective of the fake news ecosystem, composed of various types of false information, actors, and their motives.
They mainly discuss the relationship between fake news and political problems, and fake news mitigation strategies.
Zhang et al.~\cite{zhang2020overview} categorize fake news detection methods based on the features and methods used to detect fake news on a large scale.
Zhou et al.~\cite{zhou2020survey} provide a summary of fake news detection methods from four perspectives: (1) the false knowledge it carries, (2) its writing style, (3) its propagation patterns, and (4) the credibility of its source.
They also discuss building an explainable fake news detection system.
Guo et al.~\cite{guo2020future} focused on a comprehensive review of the new research trends of fake news detection; embracing novel machine learning models, aggregation of crowd wisdom, adversarial attack, and defense in detection models, and so on.
Collins et al.~\cite{collins2021trends} explored various methods of combating fake news on social media and discussed how the utility of the application of hybrid-machine learning techniques and the collective effort of humans are.
In addition, there are surveys of fake news detection methods, mainly based on the perspective of natural language processing (NLP) techniques~\cite{oshikawa2020survey,su2020motivations}, or based on the classification results from systematic reviews and meta-analysis methods~\cite{de2020approaches,elhadad2019fake,ahmed2021detecting}.
These reviews mainly summarize fake news detection methods, while there are also surveys about fact verification methods~\cite{thorne2018automated, kotonya2020explainable}.
Many other surveys on fake news exist~\cite{bondielli2019survey,dutta2019fake,miro2021misinformation,dwivedi2020survey,fernandez2018online}.
These surveys, as well as fake news tutorials~\cite{zhou2019fake,zafarani2019fake,nakov2020fact,giachanou2020battle,nakov2021fake} and technical books~\cite{shu2019detecting,shu2020disinformation}, are beneficial resources for beginners to start research on fake news.

The aforementioned surveys are mainly composed of an introduction to methods related to fake news detection and fact verification.
There are other surveys that introduce fake news datasets as follows, similar to our study:
Pierri et al.~\cite{pierri2019false} explore the datasets on social media that are used to instruct false news classification methods, in addition to recent studies that examine detection, characterization, and mitigation of false news.
Sharma et al.~\cite{sharma2019combating} provide a comprehensive summary of 23 datasets related to fake news, along with an introduction to recent fake news detection and mitigation methods.
D'Ulizia et al.~\cite{d2021fake} provide a comprehensive summary of 27 datasets related to fake news and compare between each dataset and discuss their content.
Compared to these surveys, our survey provides a large volume of summaries and an insightful discussion of future challenges and problems in dataset construction.
We expect that our survey will help readers to find a suitable dataset and construct a useful dataset for their research.

\subsection{Fact-checking sites}
When building a dataset on fake news research, determining whether news in a dataset is fake, often depends on the judgment of existing fact-checking sites.
Fact-checking sites provide the original truth-level of a piece of news based on investigation by experts. 
Real-time news is always a mixture of information, and sometimes, a binary classification result cannot fully explain the overall problem.
We can verify the authenticity of the various types of news on the Internet, thanks to fact-checking sites that inform users whether each information is true, false, or in-between.
In the subsection, we introduce useful fact-checking sites, which are also used for dataset construction.

\begin{itemize}
    \item \textbf{PolitiFact}~\cite{polsite} is an independent, non-partisan site for online fact-checking.
    It mainly rates the accuracy of claims or statements made by US political news and politicians' statements using the following classification: true, mostly true, half true, mostly false, false, and pants on fire. 
    It is a useful fact-checking site frequently used in the construction of famous datasets such as FakeNewsNet~\cite{shu2020fakenewsnet} and LIAR~\cite{liar}.

    \item \textbf{Snopes.com}~\cite{snopsite} is one of the first online fact-checking sites, which was launched in 1994. 
    It handles political and other social and topical issues.
    It mainly verifies news stories spread on social media. 
    Similar to PolitiFact, Snopes is also a functional fact-checking site frequently used in famous dataset construction such as Twitter15~\cite{twitter15} and Twitter16~\cite{twitter16}.
    
    \item \textbf{Suggest}~\cite{suggest}, also called as \textbf{GossipCop}, investigates fake news among the US entertainment and celebrity stories published in magazines and web portals.
    They rate on the basis of a 0--10 scale assigned to each article based on its authenticity, where 0 means the rumor is completely false or fictive, while 10 means the report is 100 percent true.
    
    \item \textbf{FactCheck.org}~\cite{factcheckorg} is a non-profit site for online fact-checking sites, describing itself as a ``consumer advocate for voters that aims to reduce the level of deception and confusion in US politics.'' 
    It mainly evaluates claims made by politicians in TV advertisements, debates, speeches, and so on. 

    \item \textbf{TruthOrFiction.com}~\cite{TruthOrFiction}is a non-partisan website where online users can quickly and easily get information about fake news and humorous or inspirational stories that are distributed through emails.
    It mainly rates the accuracy of these rumors based on various criteria: true, fiction, unproven, disputed, reported to be truth, reported to be fiction, truth and fiction, or truth but inaccurate details.

    \item \textbf{Hoax-Slayer}~\cite{hoaxslayer} was a privately run website for debunking false stories and internet scams and also hosted a page listing strange but true stories. 
    Various pieces of fake news have been examined, such as fake videos claiming to depict the Malaysia Airlines Flight 37~\cite{flight37}. 
    It was closed on May 31th, 2021.
    
    \item \textbf{The International Fact-Checking Network at Poynter}~\cite{Poynter} is a unit of the Poynter Institute dedicated to bringing together fact-checkers worldwide. It also sets a code of ethics for fact-checking organizations, reviews fact-checkers for compliance with its code, and then issues certification to publishers who pass the audit.
    
    \item \textbf{FullFact}~\cite{fullfact} is a British charity for online fact-checking sites. 
    It checks and corrects the news and claims that circulate on social media in the UK.
    
    \item \textbf{Fact Check Initiative Japan}~\cite{FIJ} is a non-profit organization for the promotion of Japanese fact-checking that is aimed at protecting society from mis/disinformation.

\end{itemize}

\section{Dataset of Fake News Detection}
Existing fact-checking sites and experts are engaged in judging the truthfulness of each piece of news. 
This is often time-consuming and demands large amounts of manual labor.
The ``Fake News Detection Task,” which assesses the truthfulness of a certain piece of news from news content or social media posts, has been performed by many researchers to save working hours and make the process automatic.
In recent years, with the development of deep learning models, many models have been proposed to achieve high detection performance.
A classical fake news detection model learns the textual style of fake news and then classifies them as fake on the basis of input data, such as information of the news content or social media posts, including the news~\cite{kwon2013prominent,weibo}.
It is difficult to detect fake news generated in the last few years using only textual information because the representation of fake news has become more diverse and complex.
Recent fake news detection models leverage textual information as well as rich contexts such as news publishers~\cite{yuan2020early}, user information~\cite{liu2018early}, temporal information~\cite{murayama2020fake}, and network information~\cite{shu2019beyond}.
Additionally, international competitions such as CheckThat!~\cite{checkthat2021} and Constraint~\cite{constraint_workshop} intensified the activities in proposing a novel fake news detection model.
Dataset construction, which provides resources for such research, is also an essential step in detecting fake news.
This section introduces the fake news detection dataset from the following two perspectives;
\begin{itemize}
    \item \textbf{News articles}: We introduce datasets that are utilized to detect fake news mainly from the body of the news article. 
    The style of each news article is an important feature for detection.
    
    \item \textbf{Social Media Posts}: We introduce datasets that are utilized to detect fake news mainly from social media posts related to each news. 
    User information and network information in social media, in addition to text in social media posts, are important features.

\end{itemize}

\subsection{News Articles}

\begin{table*}[t!]
    \centering
    \scriptsize
    \caption{Summary of datasets of fake news detection on news articles}
    \begin{tabular}{|l|l|c|l|l|l|l|}
        \hline
        \textbf{Dataset} & \textbf{Instances} & \textbf{Labels} & \multicolumn{1}{c|}{\textbf{Topic domain}} & \multicolumn{1}{c|}{\textbf{Raters}} & \textbf{Language} & \textbf{Year}\\ \hline
        \multirow{2}{*}{Politifact14~\cite{politifact14}} & \multirow{2}{*}{221 headlines} & \multirow{2}{*}{5} & \multirow{2}{*}{Politics, Society} & Fact-checking sites & \multirow{2}{*}{English} & \multirow{2}{*}{2014}\\ 
        & & & & (Politifact, Channel 4) & & \\ \hline
        \multirow{2}{*}{Buzzfeed\_political~\cite{benjamin2017}} & \multirow{2}{*}{71 articles} & \multirow{2}{*}{2} & \multirow{2}{*}{the 2016 US election} & \multirow{2}{*}{Buzzfeed page~\cite{buzz_trump}} & \multirow{2}{*}{English} & \multirow{2}{*}{2017}\\
        & & & & & &\\ \hline
        \multirow{2}{*}{Random\_political~\cite{benjamin2017}} & \multirow{2}{*}{225 articles} & \multirow{2}{*}{3} & \multirow{2}{*}{Politics} & \multirow{2}{*}{List of Zimdars~\cite{zimdars2016false}} & \multirow{2}{*}{English} & \multirow{2}{*}{2017}\\
        & & & & & &\\ \hline
        \multirow{2}{*}{Ahmed2017~\cite{ahmed2017detection}} & \multirow{2}{*}{25,200 articles} & \multirow{2}{*}{2} & \multirow{2}{*}{News in 2016} & Fact-checking site & \multirow{2}{*}{English} & \multirow{2}{*}{2017}\\
        & & & & (Politifact) & & \\ \hline 
        \multirow{2}{*}{LIAR~\cite{liar,liar_plus}} & \multirow{2}{*}{12,836 claims} & \multirow{2}{*}{6} & \multirow{2}{*}{-} & Fact-checking site & \multirow{2}{*}{English} & \multirow{2}{*}{2017}\\
        & & & & (Politifact) & & \\  \hline
        \multirow{2}{*}{TSHP-17\_politifact~\cite{varying}} & \multirow{2}{*}{10,483 statements} & \multirow{2}{*}{6} & \multirow{2}{*}{-} & Fact-checking site & \multirow{2}{*}{English} & \multirow{2}{*}{2017}\\
        & & & & (Politifact) & & \\ \hline
        \multirow{2}{*}{FakeNewsAMT~\cite{coling2018_automatic}} & \multirow{2}{*}{480 articles} & \multirow{2}{*}{2} & Sports, Business, Entertainment, & Generating fake news & \multirow{2}{*}{English} & \multirow{2}{*}{2018}\\
        & & & Politics, Technology, Education &  by Crowdsourcing & & \\ \hline
        \multirow{2}{*}{Celebrity~\cite{coling2018_automatic}} & \multirow{2}{*}{500 articles} & \multirow{2}{*}{2} & \multirow{2}{*}{Celebrity} & Fact-checking site & \multirow{2}{*}{English} & \multirow{2}{*}{2018}\\
        & & & & (GossipCop) & & \\ \hline
        \multirow{2}{*}{Kaggle\_UTK~\cite{kaggle_fake}} & \multirow{2}{*}{25,104 articles} & \multirow{2}{*}{2} & \multirow{2}{*}{-} & \multirow{2}{*}{-} & \multirow{2}{*}{English} & \multirow{2}{*}{2018}\\
        & & & & & &\\ \hline
        \multirow{2}{*}{MisInfoText\_Buzzfeed~\cite{torabi2019big}} & \multirow{2}{*}{1413 articles} & \multirow{2}{*}{4} & \multirow{2}{*}{-} & Fact-checking site & \multirow{2}{*}{English} & \multirow{2}{*}{2019}\\
        & & & & (Buzzfeed) & & \\ \hline
        \multirow{2}{*}{MisInfoText\_Snopes~\cite{torabi2019big}} & \multirow{2}{*}{312 articles} & \multirow{2}{*}{5} & \multirow{2}{*}{-} & Fact-checking site & \multirow{2}{*}{English} & \multirow{2}{*}{2019}\\
        & & & & (Snopes) & & \\ \hline
        \multirow{2}{*}{FA-KES~\cite{salem2019fa}} & \multirow{2}{*}{804 articles} & \multirow{2}{*}{2} & \multirow{2}{*}{Syrian War} & \multirow{2}{*}{Expert annotators} & \multirow{2}{*}{English} & \multirow{2}{*}{2019}\\
        & & & & & &\\ \hline
        \multirow{2}{*}{Spanish-v1~\cite{posadas2019detection}} & \multirow{2}{*}{971 articles} & \multirow{2}{*}{2} & Science, Sport, Politics, Society, & Fact-checking sites & \multirow{2}{*}{Spanish} & \multirow{2}{*}{2019}\\
        & & & Environment, International & (VerificadoMX, Maldito Bulo, Caza Hoax) & & \\ \hline
        \multirow{2}{*}{fauxtography~\cite{zlatkova-etal-2019-fact}} & \multirow{2}{*}{1,233 articles} & \multirow{2}{*}{2} & \multirow{2}{*}{-} & Fact-checking site & \multirow{2}{*}{English} & \multirow{2}{*}{2019}\\
        & & & & (Snopes) & & \\ \hline
        \multirow{2}{*}{Breaking!~\cite{2019breaking}} & \multirow{2}{*}{679 articles} & \multirow{2}{*}{3} & \multirow{2}{*}{2016 US election} & \multirow{2}{*}{BS Detector} & \multirow{2}{*}{English} & \multirow{2}{*}{2019}\\ 
        & & & & & &\\ \hline
        \multirow{2}{*}{TDS2020~\cite{fdr}} & \multirow{2}{*}{46,700 articles} & \multirow{2}{*}{2} & \multirow{2}{*}{-} & News Sites & \multirow{2}{*}{English} & \multirow{2}{*}{2020}\\
        & & & & (BreitBart, The Onion, InfoWars) & & \\ \hline
        \multirow{2}{*}{FakeCovid~\cite{fakecovid}} & \multirow{2}{*}{12,805 articles} & \multirow{2}{*}{2--18} & \multirow{2}{*}{COVID-19} & Fact-checking sites & \multirow{2}{*}{40 languages} & \multirow{2}{*}{2020}\\
        & & & & (Snopes, Poynter) & & \\ \hline
        \multirow{2}{*}{TrueFact\_FND~\cite{fnd2020}} & \multirow{2}{*}{6,236 articles} & \multirow{2}{*}{2} & \multirow{2}{*}{-} & \multirow{2}{*}{-} & \multirow{2}{*}{English} & \multirow{2}{*}{2020}\\  
        & & & & & &\\ \hline
        \multirow{2}{*}{Spanish-v2~\cite{posadas2019detection}} & \multirow{2}{*}{572 articles} & \multirow{2}{*}{2} & Science, Sport, Politics, Society, & Fact-checking sites & \multirow{2}{*}{Spanish} & \multirow{2}{*}{2021}\\ 
        & & & Environment, International & (VerificadoMX, Maldito Bulo, Caza Hoax) & & \\ 
    \hline
    \end{tabular}
    \label{dataset_fnda}
\end{table*}


\textbf{Politifact14}~\cite{politifact14} is one of the earliest datasets developed for fake news detection.
The paper introducing Politifact14 is also the first to suggest ``fact checking'' and assesses the truthfulness of news publishers’ statements.
The main element of the dataset, with 221 samples, is a statement (also called a headline).
Its label has a five-point scale: true, mostly true, half true, mostly false, and false.
Horne et al.~\cite{benjamin2017} constructed two types of datasets to analyze the differences between the styles of three news items: fake news, real news, and satire news.
One is \textbf{Buzzfeed\_political}, whose main topic is the 2016 US presidential election, constructed from Buzzfeed's 2016 article regarding fake election news on Facebook~\cite{buzz_trump}.
The dataset has 36 real news stories and 35 fake news stories.
The other is \textbf{Random\_political}, whose theme is political news, constructed from a list of Zimdars~\cite{zimdars2016false}.
It is composed of 75 real news stories, 75 fake news stories, and 75 satire news stories.
Since fake news on political topics are published more frequently than on other topics, a dataset focusing on political events, such as Buzzfeed\_political and Random\_political, is constructed.
For example, \textbf{TSHP-17\_politifact}~\cite{varying} is a dataset comprising individual statements made by public political figures that has been labeled according to the ratings used by Politifact (between True to Pants-on-fire (six classes)).
\textbf{Breaking!}~\cite{2019breaking} is a dataset comprising news during and before the 2016 US presidential election to implement a classification model based on linguistic features.
Articles in the dataset are divided into three categories: false, partial truth, and opinions, and they also label the questionability of each news item.

\textbf{Ahmed2017}~\cite{ahmed2017detection} consists mainly of news from 2016.
The number of samples in the dataset was 12,600 real articles obtained from Reuters.com and 12,600 articles judged to be fake based on Politifact.
\textbf{LIAR}~\cite{liar} is a large-scale dataset in fake news detection.
It is annotated using six fine-grained labels (true, mostly true, half true, mostly false, false, and pants on fire) and comprises 12,836 short statement claims from 2007 to 2016 along with information regarding the speaker, a label of credibility, the subject, the context of the statement, and so on, related to each claim in the form of metadata.
Alhindi et al. extend the LIAR dataset by automatically extracting for each claim the justification that people have provided in the fact-checking article associated with the claim~\cite{liar_plus}.
Pérez-Rosas et al.~\cite{coling2018_automatic} introduce two datasets covering seven different news domains for fake news detection and exploratory analysis to identify linguistic differences in fake and legitimate news content. 
In \textbf{FakeNewsAMT}, one dataset of \cite{coling2018_automatic}, legitimate news, the number of which is 240, is obtained from a variety of mainstream news websites such as ABC News, CNN, USA Today, The New York Times, Fox News, Bloomberg, and CNET, among others.
They do not utilize the fake news spread on the Internet as fake news in the dataset.
They asked crowd workers in Amazon Mechanical Turk to generate 240 fake news stories based on legitimate news to cover a variety of news domains.
In \textbf{Celebrity}, a dataset from \cite{coling2018_automatic}, legitimate news is obtained from online magazines such as Entertainment Weekly, People Magazine, RadarOnline, among other tabloid and entertainment-oriented publications.
Also, fake news is obtained based on the ratings given by GossipCop.
It is composed of 250 legitimate news and 250 fake news stories.
Torabi Asr et al.~\cite{torabi2019big} built two datasets of news article texts that were labeled by fact-checking websites to address the lack of data with reliable labels.
Each sample of \textbf{MisInfoText\_Buzzfeed}, one dataset of \cite{torabi2019big}, is crawled based on the labeling followed by Buzzfeed.
It comprises 1,090 mostly true news, 170 mixture of true and false news, 64 mostly false news, and 56 articles containing no factual content.
Another dataset is \textbf{MisInfoText\_Snopes}, built based on verified articles in Snopes.
The collected articles are assigned to Snopes' labels, such as fully true, mostly true, mixture of true and false, mostly false, and fully false.

Kaggle~\cite{kaggle_main}, an online community of data scientists and machine learning practitioners, has become a platform for publishing fake news datasets, despite the issue that the method of creating datasets is not clear by research paper.
\textbf{Kaggle\_UTK}~\cite{kaggle_fake} provided by Kaggle is a dataset for classifying reliable and unreliable news articles.
\textbf{TDS2020}~\cite{fdr} is a dataset that combines the Kaggle dataset comprising articles from fake news resources such as BreitBart, The Onion, and InfoWars and some mainstream web articles from CNN, BCC, The Guardian, and so on, as real news, to enhance the fake news detection model.
It comprises 24,194 fake news articles and 22,506 true news articles published after 2015.
\textbf{TrueFact\_FND}~\cite{fnd2020}, which is hosted on Kaggle, is a dataset prepared for one of the shared tasks in KDD 2020 TrueFact Workshop: Making a Credible Web for Tomorrow.
The competition, which aims to build a high-quality fake news detection model prepares the original dataset.

In addition to the above datasets, various datasets have been built for fake news detection tasks from news article contents.
\textbf{FA-KES}~\cite{salem2019fa} is a dataset constructed to focus on the specific nature of news reporting on war incidents, especially the Syrian War.
They used keywords relevant and specific to each of the events in the war, and then built a corpus of 804 news articles of news sites extoling various political positions, such as Reuters, Etilaf, SANA, Al Arabiya, The Lebanese National News Agency, Sputnik, and so on.
They labeled 426 true articles and 376 fake articles, based on whether the VDC database containing records of the casualties during the Syrian conflict and the extraction of casualty information from each news article by crowdsourcing are in agreement.
Posadas-Durá et al.~\cite{posadas2019detection} built fake news detection datasets, which are collections of Spanish news compiled from several resources on the web: \textbf{Spanish-v1} and \textbf{Spanish-v2}.
A total of 1,223 news articles in the dataset are tagged considering only two classes, true or fake, with regard to fact-checking sites for a community of Hispanic origins, such as VerificadoMX, Maldito Bulo, and Caza Hoax.
Zlatkova et al. propose a novel dataset called \textbf{fauxtography}~\cite{zlatkova-etal-2019-fact}, focusing on the relationship between fake news and images.
Each sample is provided as an image-claim pair for a new task to predict the factuality of a claim with respect to an image.
Image-claim pairs tagged as fake are gathered from a special section for image-related fact-checking, called Fauxtography in Snopes.
It comprises 641 false pairs and 592 true pairs.
Shahi et al. propose the first multilingual cross-domain dataset of 5,182 fact-checked news articles for COVID-19, collected from January 4th, 2020 to May 15th, 2020, called \textbf{FakeCovid}~\cite{fakecovid}.
They collect fact-checked articles and classify the truthfulness rate of each article, following the judgment of 92 different fact-checking websites.
The dataset, which includes posts in 40 languages from 105 countries, is utilized as a baseline for CheckThat! Task 3 in CLEF2021.
The summary of these datasets is shown in Table~\ref{dataset_fnda}.

\subsection{Social Media Posts}
Fake news detection on social media has become an important task because the development of social media has caused an increase in users catching fake news~\cite{allcott2017social}.
In addition, we can build a fake news detection model with high quality, not only by leveraging textual information but also contextual information such as user information and network information on social media.
These backgrounds activate dataset construction for fake news detection from social media.
The summary of these datasets is shown in Table~\ref{dataset_fnds}.

The performance comparison of the fake news detection model from social media posts frequently leverages some specific datasets: Twitter15, Twitter16, Twitter-ma, PHEME, and FakeNewsNet.
Ma et al. construct \textbf{Twitter15} and \textbf{Twitter16}, which are the most standard datasets for fake news detection~\cite{twitter16}, based on \cite{weibo} and \cite{twitter15}.
They find source tweets that are highly retweeted or replies related to each news item, and gather all the propagation threads on Twitter.
In addition, they assign four labels to the source tweets by referring to the labels of the events they are from: true rumors, false rumors, non-rumors, and unverified rumors.
\textbf{Twitter-ma}~\cite{weibo} is a dataset for the task of classifying rumor or non-rumors, and is composed of reported events during March-December 2015 by Snopes.
For the dataset construction, they crawl threads, which comprise 498 rumors and 494 non-rumor news from Twitter.
\textbf{PHEME}~\cite{pheme}, is a dataset collected and assigned three labels, and includes 330 threads related to nine different breaking threads, such as Prince to play in Toronto, the Ottawa shooting, and Ferguson unrest.
The dataset contains conversations on Twitter initiated by a rumor tweet.
Each tweet was grouped by story, and then either annotated based on whether the stories were true or false once confirmed, or unverified if they could not be confirmed during the collection period.
\textbf{PHEME-update}~\cite{kochkina2018all} is the extended version of the PHEME dataset~\cite{pheme}.
This dataset contains three levels of annotation. 
First, each thread is annotated as either rumor or non-rumor; second, rumors are labeled as either true, false, or unverified. Third, each tweet was annotated for stance classification through crowdsourcing. 
The number of rumors in the dataset was 2,402, with 1,067 true rumors, 638 false rumors, and 697 unverified rumors.
\textbf{FakeNewsNet}~\cite{shu2020fakenewsnet} is a dataset for the fake news detection task that contains a rich social media context and has been used in many studies.
The dataset contains articles and related tweets fact-checked by PolitiFact or GossipCop.
They retrieve 467 thousand tweets in the PolitiFact dataset, and 1.25 million in GossipCop, and label them as either real and fake.
Its strengths include the availability of rich social context information, such as original posts, response and res-shared posts of them, user profiles, followers/followees, and social networks.
However, there is the issue of a significant amount of time required to gather tweets using Twitter API, owing to the volume of the information.

In addition to these famous datasets, various researchers have created other datasets that are suitable for the proposed model or add new information.
Jiang et al. propose a dataset of 5,303 social media posts with 2,615,373 comments from multiple social media platforms such as Facebook, Twitter, and YouTube, and called \textbf{Jiang2018}, to analyze the linguistic difference between the posts related to true and false articles~\cite{jiang}.
They use 5-way scaling to rate each post following the judgment criteria of fact-checking sites: true, mostly true, half true, mostly false, and false.
\textbf{Rumor-anomaly}~\cite{tam2019anomaly} is a dataset of large social network information including posts on Twitter used for showing the effectiveness of their approach to detect rumors at the network level, following a graph-based scan approach.
It comprises four million tweets, three million users, 28,893 hashtags, and 305,115 linked articles, revolving around 1,022 rumors from May 1st, 2017 to November 1st, 2017 that contain several rumors related to the Las Vegas shooting and information published by the US administration.
Each sample is annotated following the Snopes rating.
\textbf{Fang}~\cite{fang} is a dataset that is a combination of three datasets; PHEME~\cite{pheme}, Twitter-ma~\cite{weibo} and FakeNewsNet~\cite{shu2020fakenewsnet}.
The dataset that included the stance information was utilized to evaluate the performance of their model.

Certain competitions related to fake news detection are held, and these organizers sometimes prepare a novel dataset to improve the techniques of fake news detection in social media posts.
\textbf{MediaEval\_Dataset}~\cite{image-verification} is utilized in the competition ``Verifying Multimedia Use'' in MediaEval2015 and MeidaEval2016, which is a benchmarking initiative dedicated to evaluating new algorithms for multimedia access and retrieval and attracts participants interested in multimodal approaches to multimedia.
The dataset is a set of fake and real social media posts mainly shared on Twitter to build a classifier for posts containing multimedia.
The strength of the dataset is that it contains many posts with images and videos.
\textbf{RumorEval2017}~\cite{RumorEval2017} and \textbf{RumorEval2019}~\cite{Rumoreval2019} are utilized in the workshop ``RumorEval'' in SemEval, which evaluates semantic analysis systems for exploring the nature of meaning in language.
RumorEval holds two tasks: stance classification toward rumors and veracity classification, using these datasets, RumorEval2017 and RumorEval2019, which are comprised of sourced posts with replies.
They annotated four labels (support, deny, query, or comment) for stance classification and three labels (true, false, or unverified) for veracity classification.

While almost all datasets consist of posts and comments on Twitter, owing to the convenience of Twitter API, some datasets are mainly comprised of posts on other social media platforms such as Sina Weibo, Facebook, Reddit and WhatsApp.
\textbf{RUMDECT}~\cite{weibo} and \textbf{Media\_Weibo}~\cite{jin2017multimodal} are a dataset collected from one of China's social media platforms, Sina Weibo.
Sina Weibo posts in these datasets are classified based on the judgment of Sina community management, which examines the doubtful posts reported by users and verifies them as false or real based on users’ reputation.
RUMDECT consists of 2,313 rumors and 2,351 non-rumors that build a rumor detection model.
Media\_Weibo aims to build a multimedia dataset that includes images, similar to Medieval\_Dataset, and comprises original post texts, attached images, and available social contexts including rumor and non-rumor sources.
\textbf{BuzzFace}~\cite{buzzface} and \textbf{Some-like-it-hoax}~\cite{some_like} are datasets collected from posts on Facebook.
BuzzFace is based on the BuzzFeed dataset~\cite{buzzfeed} that consists of 2,282 articles along with several of Facebook’s features (e.g., number of likes) and the assigned veracity rating.
They crawl comments and reactions related to articles in the BuzzFeed dataset on Facebook and get over 1.6 million comments using the Facebook Graph API.
Each article in BuzzFace has four categorized groups: no factual content, a mixture of true and false, mostly true, or mostly false.
Some-like-it-hoax consists of 15,500 Facebook posts in the scientific field and reactions by 909,236 users to verify whether the classification of hoaxes as real or fake is possible based on user reactions.
\textbf{Fakeddit}~\cite{fakeddit} and \textbf{Reddit\_comments}~\cite{setty2020truth} are datasets collected from threads on Reddit.
Fakeddit is a large multimodal dataset consisting of over 1 million submissions from 22 different subreddits and multiple categories of fake news from March 19, 2008, to October 24, 2019.
The dataset includes the submission title and image, comments, and various submission metadata, including the score, up-vote to down-vote ratio, and number of comments.
It is classified based on fine-grained fake news categorizations: 2-way (fake and true), 3-way (fake, a mixture of fake and true, and true), and 6-way (true, satire/parody, misleading content, imposter content, false connection, and manipulated content).
Setty et al. also proposed a Reddit-based fake news detection dataset, Reddit\_comments, comprising 12,597 threads with over 662,000 comments.
Wang et al. constructed \textbf{WeChat\_Dataset}~\cite{wefend} to test whether they can leverage user reports as weak supervision for fake news detection.
The dataset includes a large collection of news articles published via WeChat official accounts and associated user reports.
WhatsApp~\cite{whatapp} is a dataset focusing on the spread of fake news on two events, the 2018 Brazilian elections and the 2019 Indian elections on WhatsApp, where misinformation campaigns have reportedly been used.
The dataset mainly consists of fact-checked images labeled as misinformation or not-misinformation, which were searched from the WhatsApp dataset using a perceptual hashing approach.

The COVID-19 pandemic has caused people to notice that fake news in health topics, such as the human papillomavirus (HPV) vaccine~\cite{tomaszewski2021identifying} and COVID-19 vaccine\cite{apuke2021fake}, can have a social impact on people.
This is why the construction of the fake news detection dataset, focusing on a health-related topic, mainly COVID-19, is gaining speed.
Dai et al. proposed two datasets: each sample in \textbf{HealthStory} is reported by news media such as Reuters Health, and each sample in \textbf{HealthRelease} is from various institutes, including universities, research centers, and companies~\cite{dai2020ginger}.
These samples are annotated based on whether each news is fake, as per the evaluation by experts in HealthNewsReview.org.
The dataset includes wide-ranging context-based information related to the news, such as user profiles, user networks, and retweets, for analyses.
\textbf{CoAID}~\cite{cui2020coaid}, a general fake news detection dataset related to COVID-19 from social media posts includes 4,251 news and 296,000 related user engagements, ranging from December 1st, 2019 to September 1st, 2020.
\textbf{COVID-HeRA}~\cite{dharawat2020drink}, the extension of the CoAID dataset, has been constructed to flag unreliable posts based on the potential risk and severity of the statements and understanding the impact of COVID-19 misinformation in health-related decision-making.
The dataset is classified into five categories: real news/claims, not severe, possibly severe misinformation, highly severe misinformation, or refutes/rebuts misinformation.
\textbf{Constraint}~\cite{constraint}, which is used in the CONSTRAINT 2021 shared task, consists of social media posts in Twitter to identify whether it contains real or fake information.
\textbf{COVID-RUMOR}~\cite{cheng2021covid} is a COVID-19 rumor dataset used for the study of sentiment analysis and other rumor classification tasks, including stance verification of COVID-19 rumors.
A total of 6,834 samples, including 4,129 articles and 2,705 tweets, were annotated with sentiment and stance labels, in addition to veracity labels (true, false, and unverified).

These datasets mainly comprise English posts, while some datasets for other languages have been constructed for the social impact of COVID-19 as a global event.
\textbf{MM-COVID}~\cite{li2020mm} is a multilingual and multimodal dataset including 3,981 pieces of fake news content and 7,192 pieces of true news content from English, Spanish, Portuguese, Hindi, French, and Italian, verified by Snopes and Poynter.
\textbf{COVID-Alam}~\cite{alam2021fighting} is also a multilingual dataset covering Arabic and English.
Expert annotators label each post of the dataset in detail regarding five questions, including ``To what extent does the tweet appear to contain false information?'' and ``Will the tweet’s claim have an impact on or be of interest to the general public?''
\textbf{ArCOV19-Rumors}~\cite{ArCOV19-Rumors}, which is the extension of an Arabic Twitter dataset ArCOV-19~\cite{haouari2021arcov}, includes 162 verified claims and relevant tweets to these claims.
The labeling rule in ArCOV19-Rumors aims to support two kinds of misinformation detection problems over Twitter: claim-level verification, which is a two-class (fake or not) classification task for each claim and all corresponding relevant tweets, and tweet-level verification, which is also a two-class classification task given a tweet with its propagation network such as reply and re-share information.
\textbf{COVID-19-FAKES}~\cite{COVID-19-FAKES} is also an COVID-19 dataset including Arabic tweets, consisting of 3,047,255 posts collected using certain keywords and are labeled as real or misleading.
\textbf{Indic-covid}~\cite{kar2020no} is an Indian dataset that collected Hindi and Bengali tweets to detect fake news in the early stagy of COVID-19 pandemic from social media.
\textbf{CHECKED}~\cite{yang2021checked} is the first Chinese dataset on COVID-19 misinformation.
The dataset provides 2,104 verified microblogs from December 2019 to August 2020, with rich context information: 1,868,175 reposts, 1,185,702 comments, and 56,852,736 likes that reveal the spread and reaction to these verified microblogs on Weibo.

\begin{table*}[t!]
    \centering
    \scriptsize
    \caption{Summary of datasets of fake news detection on social media posts}
    \begin{tabular}{|l|l|c|l|l|l|l|l|}
        \hline
        \textbf{Dataset} & \textbf{Instances} & \textbf{Labels} & \multicolumn{1}{c|}{\textbf{Topic Domain}} & \multicolumn{1}{c|}{\textbf{Raters}} & \multicolumn{1}{c|}{\textbf{Platform}} &\textbf{Language} & \textbf{Year}\\ \hline
        \multirow{2}{*}{MediaEval\_Dataset~\cite{image-verification}} & \multirow{2}{*}{15,629 posts} & \multirow{2}{*}{2} & \multirow{2}{*}{-} & \multirow{2}{*}{-} & Twitter, Facebook, & \multirow{2}{*}{English} & \multirow{2}{*}{2015}\\
        & & & & & Blog Post & & \\ \hline
        \multirow{2}{*}{PHEME~\cite{pheme}} & \multirow{2}{*}{330 threads} & \multirow{2}{*}{3} & \multirow{2}{*}{Society, Politics} & \multirow{2}{*}{Crowdsourcing} & \multirow{2}{*}{Twitter} & \multirow{2}{*}{English} & \multirow{2}{*}{2016}\\
        & & & & & & & \\ \hline
        \multirow{2}{*}{Twitter-ma~\cite{weibo}} & \multirow{2}{*}{992 threads} & \multirow{2}{*}{2} & \multirow{2}{*}{-} & Fact-checking site & \multirow{2}{*}{Twitter} & \multirow{2}{*}{English} & \multirow{2}{*}{2016}\\
        & & & & (Snopes) & & & \\ \hline
        \multirow{2}{*}{RUMDECT~\cite{weibo}} & \multirow{2}{*}{4,664 threads} & \multirow{2}{*}{2} & \multirow{2}{*}{-} & \multirow{2}{*}{Sina community management} & \multirow{2}{*}{Weibo} & \multirow{2}{*}{Chinese} & \multirow{2}{*}{2016}\\
        & & & & & & & \\ \hline
        \multirow{2}{*}{RumorEval2017~\cite{RumorEval2017}} & \multirow{2}{*}{297 threads} & \multirow{2}{*}{3} & \multirow{2}{*}{-} & \multirow{2}{*}{PHEME~\cite{pheme}} & \multirow{2}{*}{Twitter} & \multirow{2}{*}{English} & \multirow{2}{*}{2016}\\
        & & & & & & & \\ \hline
        \multirow{2}{*}{Twitter15~\cite{twitter16}} & \multirow{2}{*}{1,478 threads} & \multirow{2}{*}{4} & \multirow{2}{*}{-} & Fact-checking sites & \multirow{2}{*}{Twitter} & \multirow{2}{*}{English} & \multirow{2}{*}{2017}\\
        & & & & (Snopes, emergent) & & & \\ \hline
        \multirow{2}{*}{Twitter16~\cite{twitter16}} & \multirow{2}{*}{818 threads} & \multirow{2}{*}{4} & \multirow{2}{*}{-} & Fact-checking sites & \multirow{2}{*}{Twitter} & \multirow{2}{*}{English} & \multirow{2}{*}{2017}\\
        & & & & (Snopes, emergent) & & & \\ \hline
        \multirow{2}{*}{BuzzFace~\cite{buzzface}} & \multirow{2}{*}{2,263 threads} & \multirow{2}{*}{4} & \multirow{2}{*}{Politics} & \multirow{2}{*}{Buzzfeed~\cite{buzzfeed}} & \multirow{2}{*}{Facebook} & \multirow{2}{*}{English} & \multirow{2}{*}{2017}\\
        & & & & & & & \\ \hline
        \multirow{2}{*}{Some-like-it-hoax~\cite{some_like}} & \multirow{2}{*}{15,500 posts} & \multirow{2}{*}{2} & \multirow{2}{*}{Science} & \multirow{2}{*}{\cite{bessi2015science}} & \multirow{2}{*}{Facebook} & \multirow{2}{*}{English} & \multirow{2}{*}{2017}\\
        & & & & & & & \\ \hline
        \multirow{2}{*}{Media\_Weibo~\cite{jin2017multimodal}} & \multirow{2}{*}{9,528 posts} & \multirow{2}{*}{2} & \multirow{2}{*}{-} & \multirow{2}{*}{Sina community management} & \multirow{2}{*}{Weibo} & \multirow{2}{*}{Chinese} & \multirow{2}{*}{2017}\\
        & & & & & & & \\ \hline
        \multirow{2}{*}{PHEME-update~\cite{kochkina2018all}} & \multirow{2}{*}{6,425 threads} & \multirow{2}{*}{3} & \multirow{2}{*}{Society, Politics} & \multirow{2}{*}{PHEME~\cite{pheme}} & \multirow{2}{*}{Twitter} & \multirow{2}{*}{English} & \multirow{2}{*}{2018}\\
        & & & & & & & \\ \hline
        \multirow{2}{*}{FakeNewsNet~\cite{shu2020fakenewsnet}} & \multirow{2}{*}{23,921 news} & \multirow{2}{*}{2} & \multirow{2}{*}{Politics, Celebrity} & Fact-checking sites & \multirow{2}{*}{Twitter} & \multirow{2}{*}{English} & \multirow{2}{*}{2018}\\
        & & & & (Politifact, GossipCop) & & & \\ \hline
        \multirow{2}{*}{Jiang2018~\cite{jiang}} & \multirow{2}{*}{5,303 posts} & \multirow{2}{*}{5} & \multirow{2}{*}{-} & Fact-checking sites & Twitter, Youtube, & \multirow{2}{*}{English} & \multirow{2}{*}{2018}\\
        & & & & (Politcact, Snopes) & Facebook & & \\ \hline
        \multirow{2}{*}{RumorEval2019~\cite{Rumoreval2019}} & \multirow{2}{*}{446 threads} & \multirow{2}{*}{3} & \multirow{2}{*}{Natural disaster} & Fact-checking sites (Politcact, Snopes) & \multirow{2}{*}{Twitter, Reddit} & \multirow{2}{*}{English} & \multirow{2}{*}{2019}\\
        & & & & (Politcact, Snopes) & & & \\ \hline
        \multirow{2}{*}{Rumor-anomaly~\cite{tam2019anomaly}} & \multirow{2}{*}{1,022 threads} & \multirow{2}{*}{6} & Politics, Fraud \& Scam, & Fact-checking site & \multirow{2}{*}{Twitter} & \multirow{2}{*}{English} & \multirow{2}{*}{2019}\\
        & & & Crime, Science, etc. & (Snopes) & & & \\ \hline
        \multirow{2}{*}{WeChat\_Dataset~\cite{wefend}} & \multirow{2}{*}{4,180 news} & \multirow{2}{*}{2} & \multirow{2}{*}{-} & \multirow{2}{*}{WeChat} & \multirow{2}{*}{WeChat} & \multirow{2}{*}{English} & \multirow{2}{*}{2020}\\
        & & & & & & & \\ \hline
        \multirow{2}{*}{Fang~\cite{fang}} & \multirow{2}{*}{1,054 threads} & \multirow{2}{*}{2} & \multirow{2}{*}{-} & PHEME~\cite{pheme}, Twitter-ma~\cite{weibo}, & \multirow{2}{*}{Twitter} & \multirow{2}{*}{English} & \multirow{2}{*}{2020}\\
        & & & & FakeNewsNet~\cite{shu2020fakenewsnet} & & & \\ \hline
        \multirow{2}{*}{WhatsApp~\cite{whatapp}} & \multirow{2}{*}{3,083 images} & \multirow{2}{*}{2} & Brazilian elections, & Fact-checking sites & \multirow{2}{*}{WhatsApp} & \multirow{2}{*}{-} & \multirow{2}{*}{2020}\\
        & & & Indian elections & (aosfatos.org, boomlive.in, e-farsas, etc.) & & & \\ \hline
        \multirow{2}{*}{Fakeddit~\cite{fakeddit}} & \multirow{2}{*}{1,063,106 posts} & \multirow{2}{*}{2,3,6} & \multirow{2}{*}{-} & \multirow{2}{*}{Expert annotators} & \multirow{2}{*}{Reddit} & \multirow{2}{*}{English} & \multirow{2}{*}{2020}\\
        & & & & & & & \\ \hline
        \multirow{2}{*}{Reddit\_comments~\cite{setty2020truth}} & \multirow{2}{*}{12,597 claims} & \multirow{2}{*}{2} & \multirow{2}{*}{-} & Fact-checkiing sites & \multirow{2}{*}{Reddit} & \multirow{2}{*}{English} & \multirow{2}{*}{2020}\\
        & & & & (Snopes, Politifact, emergent) & & & \\ \hline
        \multirow{2}{*}{HealthStory~\cite{dai2020ginger}} & \multirow{2}{*}{1,690 threads} & \multirow{2}{*}{2} & \multirow{2}{*}{Health} & \multirow{2}{*}{HealthNewsReview} & \multirow{2}{*}{Twitter} & \multirow{2}{*}{English} & \multirow{2}{*}{2020}\\
        & & & & & & & \\ \hline
        \multirow{2}{*}{HealthRelease~\cite{dai2020ginger}} & \multirow{2}{*}{606 threads} & \multirow{2}{*}{2} & \multirow{2}{*}{Health} & \multirow{2}{*}{HealthNewsReview} & \multirow{2}{*}{Twitter} & \multirow{2}{*}{English} & \multirow{2}{*}{2020}\\
        & & & & & & & \\ \hline
        \multirow{2}{*}{CoAID~\cite{cui2020coaid}} & \multirow{2}{*}{4,251 threads} & \multirow{2}{*}{2} & \multirow{2}{*}{COVID-19} & Fact-checking sites & \multirow{2}{*}{Twitter} & \multirow{2}{*}{English} & \multirow{2}{*}{2020}\\
        & & & & (Politifact, FactCheck.org, etc.) & & & \\ \hline
        \multirow{2}{*}{COVID-HeRA~\cite{dharawat2020drink}} & \multirow{2}{*}{61,286 posts} & \multirow{2}{*}{5} & \multirow{2}{*}{COVID-19} & \multirow{2}{*}{CoAID~\cite{cui2020coaid}, Expert annotators} & \multirow{2}{*}{Twitter} & \multirow{2}{*}{English} & \multirow{2}{*}{2020}\\
        & & & & & & & \\ \hline
        \multirow{2}{*}{ArCOV19-Rumors~\cite{ArCOV19-Rumors}} & \multirow{2}{*}{162 threads} & \multirow{2}{*}{2} & \multirow{2}{*}{COVID-19} & Fact-checking sites & \multirow{2}{*}{Twitter} & \multirow{2}{*}{Arabic} & \multirow{2}{*}{2020}\\
        & & & & (Fatabyyano, Misbar) & & & \\ \hline
        \multirow{3}{*}{MM-COVID~\cite{li2020mm}} & \multirow{3}{*}{11,173 threads} & \multirow{3}{*}{2} & \multirow{3}{*}{COVID-19} & \multirow{2}{*}{Fact-checking sites} & \multirow{3}{*}{Twitter} & English, Spanish, & \multirow{3}{*}{2020}\\
        & & & & \multirow{2}{*}{(Snopes, Poynter)} & & Portuguese, Hindi, & \\
        & & & & & & French, Italian & \\ \hline
        \multirow{2}{*}{Constraint~\cite{constraint}} & \multirow{2}{*}{10,700 posts} & \multirow{2}{*}{2} & \multirow{2}{*}{COVID-19} & Fact-checking sites & \multirow{2}{*}{Twitter} & \multirow{2}{*}{English} & \multirow{2}{*}{2020}\\
        & & & & (Politifact, Snopes) & & & \\ \hline
        \multirow{2}{*}{Indic-covid~\cite{kar2020no}} & \multirow{2}{*}{1,438 posts} & \multirow{2}{*}{2} & \multirow{2}{*}{COVID-19} & \multirow{2}{*}{Expert annotators} & \multirow{2}{*}{Twitter} & Bengali, & \multirow{2}{*}{2020}\\
        & & & & & & Hindi & \\ \hline
        \multirow{2}{*}{COVID-19-FAKES~\cite{COVID-19-FAKES}} & \multirow{2}{*}{3,047,255 posts} & \multirow{2}{*}{2} & \multirow{2}{*}{COVID-19} & WHO, UN, & \multirow{2}{*}{Twitter} & Arabic, & \multirow{2}{*}{2020}\\
        & & & & UNICEF & & English & \\ \hline
        \multirow{2}{*}{CHECKED~\cite{yang2021checked}} & \multirow{2}{*}{2,104 threads} & \multirow{2}{*}{2} & \multirow{2}{*}{COVID-19} & \multirow{2}{*}{Sina community management} & \multirow{2}{*}{Weibo} & \multirow{2}{*}{Chinese} & \multirow{2}{*}{2021}\\
        & & & & & & & \\ \hline
        \multirow{2}{*}{COVID-Alam~\cite{alam2021fighting}}& \multirow{2}{*}{722 tweets} & \multirow{2}{*}{5} & \multirow{2}{*}{COVID-19} & \multirow{2}{*}{Expert annotators} & \multirow{2}{*}{Twitter} & \multirow{2}{*}{English, Arabic} & \multirow{2}{*}{2021}\\
        & & & & & &  & \\ \hline
        \multirow{2}{*}{COVID-RUMOR~\cite{cheng2021covid}} & \multirow{2}{*}{2,705 posts} & \multirow{2}{*}{2} & \multirow{2}{*}{COVID-19} & Fact-checking sites & Twitter, & \multirow{2}{*}{English} & \multirow{2}{*}{2021}\\
        & & & & (Snopes, Politifact, Boomlive) & Websites & & \\ \hline
    \end{tabular}
    \label{dataset_fnds}
\end{table*}


\section{Dataset of Fact Verification}
There are works by journalists or experts to determine whether a specific claim is correct based on a given evidence in the process of fact-checking.
The attempt to automatically process these judgments by NLP technology is called as ``Fact Verification'' task.
Unlike the fake news detection task, which is to classify whether the news is correct without being (explicitly) given the evidence from the writing style and network information of social media posts related to it, a fact-verification task helps to make the decision as to whether a claim is correct, specifically, when explicitly given the evidence.
Some studies such as ~\cite{thorne2018automated} handle fake news detection on social media tasks in our survey as part of a fact-verification task, because they regard text information, network information, and aggregate information on the posts on social networks as evidence, but we treat these tasks differently.
A model for a fact-verification task classifies whether each claim is correct from a claim and the evidence given as input data.
Sets of a claim and given evidence are generally utilized as input data for the classification model of a fact-verification task.
A claim is mainly given in a single sentence, like the headline of an article.
There are many different styles of given evidence: small knowledge graphs ~\cite{vlachos2015identification}, table data~\cite{2019TabFactA}, web data~\cite{bast2017overview}, and text data~\cite{liar}.
For example, the Fact Extraction and VERification (FEVER) task~\cite{fever1}, which is the biggest competition for fact verification, requires combining information from multiple documents and sentences for fact-checking. 
Unlike the aforementioned works, which use text as evidence, the evidence is not given but must be retrieved from Wikipedia, a large corpus of encyclopedic articles.
In other words, the FEVER task contains not only veracity prediction but also document retrieval and evidence selection tasks.

The fact-verification task is important and practical in the process of the fact-checking process; however, the number of datasets on the fact-verification task is smaller than that on the fake news detection task.
This is because the relative scarcity of resources is designed explicitly for fact-checking, which highlights the difficulties in capturing and formalizing the task.
In addition, fact verification is similar to these tasks such as ``textual entailment'' or ``natural language inference'' in NLP field.
Thorne et al. discussed a key difference in these tasks, where ``each claim typically consists of a single sentence, while each claim is retrieved from a large set of documents in fact verification.''
Some datasets with awareness of fact-checking, such as for example, FEVER, fit the definition; however, some provide the evidence data without retrieval.
It is difficult to divide each dataset based on whether it is for text entailment or verification.
This section introduces the dataset meeting the requirements as below: (1) it is constructed to solve the task of classifying a class of given claims based on the evidence, we call ``fact verification task,'' (2) the dataset is constructed on the awareness of the problems related to fake news such as fact-checking.
Numerical~\cite{vlachos2015identification}, TabFacts~\cite{2019TabFactA}, and InfoTABS~\cite{gupta-etal-2020-infotabs}, and so on, are famous datasets for text entailment and natural language inference tasks, but we do not introduce these datasets because they do not present fact-checking problems in their papers.
The summary of these datasets is shown in Table~\ref{dataset_fv}.

Popat et al.~\cite{popat2016credibility} first addressed the task of assessing the credibility of arbitrary claims made in natural language text to solve the problems of fake claims and then created two datasets, \textbf{Snopes\_credibility} and \textbf{Wikipedia\_credibility}.
Snopes\_credibility consists of validated claims and 30 web articles related to each claim.
Wikipedia\_credibility, which is constructed to test the generalization performance of models trained on Snopes\_credibility, consists of fake articles and web articles related to each claim based on the list of fictitious people and lists of hoaxes.
\textbf{DeClarE\_politifact}~\cite{popat-etal-2018-declare} consists of claims from Politifact published before December 2017 and websites, that is created by the same method as Snopes\_credibility~\cite{popat2017truth}.
\textbf{UKPSnopes}~\cite{richly} is constructed for the purpose of being a substantially sized mixed-domain corpus with annotations for several tasks, document retrieval, evidence extraction, stance detection, and claim validation throughout the fact-checking.
Components in UKPSnopes, claims, evidence texts, and original documents are extracted from Snopes.
Each claim has 5-way labels and each claim-evidence text pair has 3-way labels (support, refute, and no stance).
\textbf{MultiFC}~\cite{multifc} consists of 36,534 paired with textual sources and rich metadata, extracted from 24 fact-checking sites, labeled for veracity following these sites with evidence pages to verify the claims and the context in which they occurred.

FakeNewsChallenge (FNC), one of the competitions, explores how artificial intelligence technologies, particularly machine learning and natural language processing, might be leveraged to combat the fake news problem and provide a novel dataset, FNC\_dataset.
\textbf{FNC\_dataset}~\cite{fnc} is developed as the helpful first step towards identifying fake news by understanding what other news organizations are saying about the topic, based on Emergent.
It is composed of a headline and body text in each sample, and the task is to estimate the stance of a body text from a news article relative to a headline. 
Specifically, the body text may agree, disagree, discuss, or be unrelated to the headline.
They are called as stance detection task in these papers.
Our survey treats this task as part of fact verification, since the estimated stance information of each body text is closely linked to the verification of the claim.
\textbf{Emergent}~\cite{ferreira-vlachos-2016-emergent}, from which FNC\_dataset is constructed, is a dataset derived from a digital journalism project for rumor debunking.
The dataset contains 300 rumored claims with their veracity labels (true, false, or unverified) and 2,595 associated news articles with the labels, which shows whether its stance is for, against, or observing the claim.

There are other competitions and workshops on fact verification, besides FNC, among which FEVER and CheckThat! are famous.
FEVER is a workshop that promotes fact-verification research and also provides some novel datasets for the shared task.
\textbf{FEVER}~\cite{fever1} dataset for the first FEVER shared task consists of 185,445 claims, generated by altering sentences extracted from Wikipedia, and pre-processed Wikipedia data.
Each claim has a three-way classification label about whether a claim is supported or refused by Wikipedia, or Wikipedia does not have enough information about the claim  (NotEnough Info); in the case of the first two labels, the sentences form the evidence to reach the verdict.
Participants in the competition need to develop the pipeline by first identifying relevant documents from Wikipedia data, then selecting sentences forming the evidence from the documents, and finally classifying the claim into three-way labels.
In the \textbf{FEVER 2.0}~\cite{fever2} dataset for the second FEVER shared task, 1,174 new claims were added as an extension of the FEVER dataset.
\textbf{FEVEROUS}~\cite{Aly21Feverous}, for the fourth FEVER shared task, is constructed for evaluating the ability of system to verify information using unstructured and structured evidence from Wikipedia.
This is the same as the FEVER dataset in terms of assigned kinds of labels and each claim generated from Wikipedia, while the difference is that FEVEROUS covers not only natural language sentences but also structured data such as table data.

The CheckThat! Lab at the Conference and Labs of the Evaluation Forum (CLEF) prepared various datasets to fight misinformation and disinformation in social media, political debates, and news.
The competition provides multiple tasks in which the process of fact-checking, such as, for example, finding check-worthy claims and selecting evidence sentences in the article, are divided.
One of their tasks corresponds to the fact verification task.
\textbf{CT-FCC-18}~\cite{nakov2018overview} in CheckThat! 2018, in which each sample is in English or Arabic, is a dataset for the classification of whether the given claim is likely to be true, false, or unsure of its factuality among the three labels using a search system as evidence.
\textbf{CT19-T2}\cite{elsayed2019overview} in CheckThat! 2019 and \textbf{CT20-Arabic}~\cite{barron2020overview} in CheckThat! 2020 are also Arabic datasets for classifying whether the given claim is likely to be true or false, when considering web pages as evidence, with other subtasks such as ``Rerank search results,'' ``Classify search results,'' and ``Classify passages from useful pages.''

Similar to CheckThat!, in which datasets are made for fake news in Arabic as well as the English-speaking public, there are other datasets that also target Arabic language text.
For example, \textbf{Arabic\_corpus}~\cite{2018integrating} and \textbf{Arabic\_stance}~\cite{khouja-2020-stance} were developed for claim verification and textual entailment recognition tasks.
Arabic\_corpus contains 422 claims labeled for factuality, true or false, and 3,042 web pages, where each claim-document pair is labeled for a stance similar to FakeNewsChallgnge: agree, disagree, discuss, or unrelated.
Arabic\_stance consists of 4,547 fake and true claims generated from the Arabic News Texts (ANT) corpus~\cite{chouigui2017ant}, similar to the FEVER dataset, and claim pairs labeled for stance, disagree, agree, or other.

Fact verification datasets constructed in languages other than English and Arabic are as follows.
\textbf{DAST}~\cite{lillie2019joint} is a Danish fact verification dataset on Reddit.
The dataset is built to detect the stance of submission posts to source comments, which they categorize as support, deny, query, or comment, following RumorEval2017. 
\textbf{DANFEVER}~\cite{norregaard-derczynski-2021-danfever} is also a Danish fact verification dataset, each sample in which is labeled as supported, refused, or NotEnough Info on the same methodology of FEVER dataset.
\textbf{Croatian}~\cite{bosnjak-karan-2019-data} is composed of Croatian news articles and comments related to them.
Each comment is annotated with three categories: whether it contains a verified claim (verifiable, non-verifiable, Spam, or Non-Claim), whether its stance is in a support or disapprove, and whether the commentator presents their comment as positive or negative.

Recently, domain specific datasets are developing increasingly.
For example, \textbf{PUBHEALTH}~\cite{publichealth}, consisting of 11,832 claims labeled for 4-way classification (true, false, mixture, unproven), focuses on the problem setting of public health, including biomedical subjects (e.g., infectious diseases, stem cell research) and government healthcare policy (e.g., abortion, mental health, and women's health).
\textbf{SCIFACT}~\cite{scifact} is a dataset for a new task of verifying the veracity of scientific claims extracting from S2ORC~\cite{s2orc}, a publicly available corpus of millions of scientific articles, based on abstracts from the research literature as evidence.
There is a motivation to address the problem of increased doubtful and unverified scientific information due to the COVID-19 pandemic.
It contains 1,409 pairs of abstracts and scientific claims annotated with three labels: supports, refutes, or no info.
Lee et al.~\cite{prexity} propose two datasets, whose topics are science and politics during COVID-19: \textbf{COVID-19-Scientific} and \textbf{COVID-19-Politifact}, for building an unsupervised fake news detection model.
COVID-19-Scientific is constructed by collecting COVID-19-related myths and scientific truths verified by reliable sources such as the CDC and WHO.
COVID-19-Politifact is also constructed by collecting COVID-19-related non-scientific and political claims extracted from Politifact. 
Both datasets have two labels: fake or true.
\textbf{COVIDLies}~\cite{hossain-etal-2020-covidlies} is composed of pairs of the tweet and misconception related to COVID-19 for the task of classifying their relationship into agree (tweet is a positive expression of the misconception), disagree (tweet contradicts with the misconception), or no stance as stance detection.
\textbf{HoVer}~\cite{jiang2020hover} is built on the Wikipedia corpus for the many-hop fact verification task that we need informational reference from multiple sources.
While most fact-verification datasets are limited by the number of reasoning steps and the word overlapping between the claim and all evidences, HoVer requires multi-hop reasoning based on evidence from as many as four English Wikipedia articles.
Each pair of claims and evidence in HoVer, extracted from the HotpotQA dataset and mutated, is labeled as either supported, refuted, or not enough info.

\begin{table*}[t!]
    \centering
    \scriptsize
    \caption{Summary of datasets of fact verification}
    \begin{tabular}{|l|l|c|l|l|l|l|l|l|}
        \hline
        \textbf{Dataset} & \textbf{Instances} & \textbf{Labels} & \multicolumn{1}{c|}{\textbf{Topic domain}} & \multicolumn{1}{c|}{\textbf{Raters}} & \multicolumn{1}{c|}{\textbf{Original data}} & \multicolumn{1}{c|}{\textbf{Main components}} &\textbf{Language} & \textbf{Year}\\ \hline
        \multirow{2}{*}{Rumor-has-it~\cite{qazvinian-etal-2011-rumor}} & \multirow{2}{*}{10,417 posts} & \multirow{2}{*}{3} & \multirow{2}{*}{-} & \multirow{2}{*}{Expert annotators} & \multirow{2}{*}{Twitter} & \multirow{2}{*}{Tweet} & \multirow{2}{*}{English} & \multirow{2}{*}{2011}\\
        & & & & & & & & \\ \hline
        \multirow{2}{*}{Snopes\_credibility~\cite{popat2016credibility, popat2017truth}} & \multirow{2}{*}{4,856 claims} & \multirow{2}{*}{2} & \multirow{2}{*}{-} & \multirow{2}{*}{Snopes} & \multirow{2}{*}{Web data} & Claim, & \multirow{2}{*}{English} & \multirow{2}{*}{2016}\\
        & & & & & & Search page & & \\ \hline
        \multirow{2}{*}{Wikipedia\_credibility~\cite{popat2016credibility}} & \multirow{2}{*}{157 claims} & \multirow{2}{*}{-} & hoax, & \multirow{2}{*}{Wikipedia} & \multirow{2}{*}{Wikipedia} & Claim,  & \multirow{2}{*}{English} & \multirow{2}{*}{2016}\\
        & & & Fictious people & & & Search page & & \\ \hline
        \multirow{2}{*}{Emergent~\cite{ferreira-vlachos-2016-emergent}} & \multirow{2}{*}{300 claims} & \multirow{2}{*}{3} & \multirow{2}{*}{-} & Snopes, & \multirow{2}{*}{Web data} & Claim, & \multirow{2}{*}{English} & \multirow{2}{*}{2016}\\
        & & & & Hoaxalizer & & News articles & & \\ \hline
        \multirow{2}{*}{FNC\_dataset~\cite{fnc}} & \multirow{2}{*}{2,587 claims} & \multirow{2}{*}{4} & \multirow{2}{*}{-} & \multirow{2}{*}{Emergent~\cite{ferreira-vlachos-2016-emergent}} & \multirow{2}{*}{Poltifact} & Claim, & \multirow{2}{*}{English} & \multirow{2}{*}{2017}\\
        & & & & & & Body text & & \\ \hline
        \multirow{2}{*}{DeClarE\_politifact~\cite{popat-etal-2018-declare}} & \multirow{2}{*}{3,569 claims} & \multirow{2}{*}{2} & \multirow{2}{*}{-} & \multirow{2}{*}{Politifact} & \multirow{2}{*}{Politifact} & claim, meta data & \multirow{2}{*}{English} & \multirow{2}{*}{2018}\\
        & & & & & & Meta data & & \\ \hline
        \multirow{2}{*}{FEVER~\cite{fever1}} & \multirow{2}{*}{185,445 claims} & \multirow{2}{*}{3} & \multirow{2}{*}{Wikipedia} & \multirow{2}{*}{Expert annotators} & \multirow{2}{*}{Wikipedia} & Claim, & \multirow{2}{*}{English} & \multirow{2}{*}{2018}\\
        & & & & & & Wikipedia data & & \\ \hline
        \multirow{2}{*}{FEVER 2.0~\cite{fever2}} & \multirow{2}{*}{1,174 claims} & \multirow{2}{*}{3} & \multirow{2}{*}{Wikipedia} & \multirow{2}{*}{Expert annotators} & \multirow{2}{*}{Wikipedia} & \multirow{2}{*}{Claim} & \multirow{2}{*}{English} & \multirow{2}{*}{2018}\\
        & & & & & & & & \\ \hline
        \multirow{2}{*}{CT-FCC-18~\cite{nakov2018overview}} &  \multirow{2}{*}{150 claims} &  \multirow{2}{*}{3} & Politics, & Snopes, &  \multirow{2}{*}{Web data} &  \multirow{2}{*}{Claim} & English, &  \multirow{2}{*}{2018}\\
        & & & Arab-related news & FactCheck.org & & & Arabic & \\ \hline
        \multirow{2}{*}{Arabic\_corpus~\cite{2018integrating}} & \multirow{2}{*}{429 claims} & \multirow{2}{*}{2} & \multirow{2}{*}{Arab-related news} & \multirow{2}{*}{VERIFY} & \multirow{2}{*}{Web data} & Claim, & \multirow{2}{*}{Arabic} & \multirow{2}{*}{2018}\\
        & & & & & & Web page & & \\ \hline
        \multirow{2}{*}{UKPSnopes~\cite{richly}} & \multirow{2}{*}{6,422 claims} & \multirow{2}{*}{5} & \multirow{2}{*}{-} & Snopes, & \multirow{2}{*}{Snopes} & Claim, Web page, & \multirow{2}{*}{English} & \multirow{2}{*}{2019}\\
        & & & & Expert annotators & & Evidence text & & \\ \hline
        \multirow{2}{*}{MultiFC~\cite{multifc}} & \multirow{2}{*}{36,534 claims} & \multirow{2}{*}{2--40} & \multirow{2}{*}{-} & \multirow{2}{*}{-} & \multirow{2}{*}{Fact-checking sites} & Claim, & \multirow{2}{*}{English} & \multirow{2}{*}{2019}\\
        & & & & & & Search page & & \\ \hline
        \multirow{2}{*}{DAST~\cite{lillie2019joint}} & \multirow{2}{*}{3,007 posts} & \multirow{2}{*}{4} & \multirow{2}{*}{-} & \multirow{2}{*}{Expert annotators} & \multirow{2}{*}{Reddit} & Source comments, & \multirow{2}{*}{Danish} & \multirow{2}{*}{2019}\\
        & & & & & & Submission posts & & \\ \hline
        \multirow{2}{*}{Croatian~\cite{bosnjak-karan-2019-data}} & \multirow{2}{*}{904 comments} & \multirow{2}{*}{4} & \multirow{2}{*}{-} & \multirow{2}{*}{Expert annotators} & \multirow{2}{*}{24 sata} & News articles, & \multirow{2}{*}{Croatian} & \multirow{2}{*}{2019}\\
        & & & & & & News comments & & \\ \hline
        \multirow{2}{*}{CT19-T2~\cite{elsayed2019overview}} & \multirow{2}{*}{69 claims} & \multirow{2}{*}{2} & \multirow{2}{*}{Arab-related news} & \multirow{2}{*}{Expert annotators} & \multirow{2}{*}{Web data} & Claim, & \multirow{2}{*}{Arabic} & \multirow{2}{*}{2019}\\
        & & & & & & Web page & & \\ \hline
        \multirow{2}{*}{CT20-Arabic~\cite{barron2020overview}} & \multirow{2}{*}{165 claims} & \multirow{2}{*}{2} & \multirow{2}{*}{Arab-related news} & \multirow{2}{*}{Expert annotators} & \multirow{2}{*}{Twitter} & Claim, & \multirow{2}{*}{Arabic} & \multirow{2}{*}{2020}\\
        & & & & & & Web page & & \\ \hline
        \multirow{2}{*}{Arabic\_Stance~\cite{khouja-2020-stance}} & \multirow{2}{*}{4,547 claims} & \multirow{2}{*}{2} & \multirow{2}{*}{Arab-related news} & \multirow{2}{*}{Expert annotators} & \multirow{2}{*}{ANT corpus~\cite{chouigui2017ant}} & Claim, & \multirow{2}{*}{Arabic} & \multirow{2}{*}{2020}\\
        & & & & & & News articles & & \\ \hline
        \multirow{2}{*}{PUBHEALTH~\cite{publichealth}} & \multirow{2}{*}{11,832 claims} & \multirow{2}{*}{4} & \multirow{2}{*}{Health} & Fact-checking sites & \multirow{2}{*}{Web data} & Claim, News articles, & \multirow{2}{*}{English} & \multirow{2}{*}{2020}\\
        & & & & (Snopes, Politifact, etc.) & & Explanation texts & & \\ \hline
        \multirow{2}{*}{COVID-19-Scientific~\cite{prexity}} & \multirow{2}{*}{142 claims} & \multirow{2}{*}{2} & \multirow{2}{*}{Science in COVID-19} & MedicalNewsToday, & \multirow{2}{*}{CORD-19~\cite{cord-19}} & \multirow{2}{*}{Claim} & \multirow{2}{*}{English} & \multirow{2}{*}{2020}\\
        & & & & CDC, WHO & & & & \\ \hline
        \multirow{2}{*}{COVID-19-Politifact~\cite{prexity}} & \multirow{2}{*}{340 claims} & \multirow{2}{*}{2} & \multirow{2}{*}{Politics in COVID-19} & \multirow{2}{*}{Politifact} & \multirow{2}{*}{Politifact} & \multirow{2}{*}{Claim} & \multirow{2}{*}{English} & \multirow{2}{*}{2020}\\
        & & & & & & & & \\ \hline
        \multirow{2}{*}{COVIDLies~\cite{hossain-etal-2020-covidlies}} & \multirow{2}{*}{6,761 posts} & \multirow{2}{*}{3} & \multirow{2}{*}{COVID-19} & \multirow{2}{*}{Expert annotators} & \multirow{2}{*}{Twitter} & Tweets, & \multirow{2}{*}{English} & \multirow{2}{*}{2020}\\
        & & & & & & Misconception & & \\ \hline
        \multirow{2}{*}{SCIFACT~\cite{scifact}}& \multirow{2}{*}{1,490 claims} & \multirow{2}{*}{3} & \multirow{2}{*}{Scientific papers} & \multirow{2}{*}{Expert annotators} & \multirow{2}{*}{S2ORC~\cite{s2orc}} & Claim, & \multirow{2}{*}{English} & \multirow{2}{*}{2020}\\
        & & & & & & Research abstracts & & \\ \hline        
        \multirow{2}{*}{HoVer~\cite{jiang2020hover}} & \multirow{2}{*}{26,171 claims} & \multirow{2}{*}{3} & \multirow{2}{*}{Wikipedia} & \multirow{2}{*}{Expert annotators} & \multirow{2}{*}{HotpotQA dataset~\cite{yang2018hotpotqa}} & Claim, & \multirow{2}{*}{English} & \multirow{2}{*}{2020}\\
        & & & & & & Wikipedia data & & \\ \hline   
        \multirow{2}{*}{FEVEROUS~\cite{Aly21Feverous}} & \multirow{2}{*}{87,062 claims} & \multirow{2}{*}{3} & \multirow{2}{*}{Wikipedia} & \multirow{2}{*}{Expert annotators} & \multirow{2}{*}{Wikipedia} & Claim, & \multirow{2}{*}{English} & \multirow{2}{*}{2021}\\
        & & & & & & Wikipedia data & & \\ \hline   
        \multirow{2}{*}{DANFEVER~\cite{norregaard-derczynski-2021-danfever}} & \multirow{2}{*}{6,407 claims} & \multirow{2}{*}{3} & \multirow{2}{*}{Wikipedia} & \multirow{2}{*}{Expert annotators} & Wikipedia, & Claim, & \multirow{2}{*}{Danish} & \multirow{2}{*}{2021}\\
        & & & & & Den Store Danske & Wikipedia data & & \\ \hline   
    \end{tabular}
    \label{dataset_fv}
\end{table*}

\section{Other datasets related to fake news}
Various datasets have been constructed for fake news detection and fact verification, as they are major tasks, and there are other tasks related to fake news.
This section introduces datasets related to other tasks, such as satire detection, check worthy, news media credibility prediction, and so on.
A summary of these datasets published before 2019 is shown in Table~\ref{dataset_2019} and those published after 2020 in Table~\ref{dataset_2020}.

\subsection{Satire detection}
The task of detecting satire articles, which mimic true news articles, incorporating irony and non-sequitur, in an attempt to provide humorous insight, was proposed earlier than the fake news detection task.
\textbf{Satire}~\cite{acl2009_satire} is the first satire corpus consisting of 4,000 real news articles and 233 satire news articles
\textbf{SatiricalCues}~\cite{rubin2016fake} is constructed for illustrating the unique feature of satire news in contrast with their real news.
It is composed of 180 real news items and 180 satirical news items in four domains (civils, science, business, soft news). 
\textbf{Fake.vs.Satire}~\cite{fake_satire} is composed of 238 fake news and 203 satire news that are constructed for the difficult task of  identifying satirical news among false news.
They mention that, to combat fake news on a web scale, it is important to distinguish fake news from satirical news without deceptive intent.

\subsection{News (Media) Credibility}
Estimating and assessing news media's credibility is a task in which credibility is often defined in terms of quality and believability.
News articles from unreliable news media do not necessarily have the characteristics of fake news.
However, some research (arbitrarily) regard news articles from unreliable sources as fake news because fake news frequently comes from either fake news websites that only publish hoaxes or from hyperpartisan websites that present themselves as publishing real news ~\cite{buzzfeed}.
Against this background, some datasets have been constructed and annotated in terms of the reliability of each news media that publishes many articles and their characteristics.

Datasets assigned the credibility of news media or news articles include the following.
Mukherjee et al. provided a dataset called \textbf{NewsTrust\_dataset}, composed of news articles reviewed in NewsTrust, which made available the ground-truth ratings for credibility analysis of news articles~\cite{cikm2015}.
They utilized the dataset to model the joint interaction to identify highly credible news articles and truth-worthy news.
\textbf{ReCOVery}~\cite{zhou2020recovery} is a repository of 2,029 news articles and 140,820 tweets related to them on COVID-19 published from 60 news media identified with extremely high or low levels of credibility.
The dataset was designed and constructed to facilitate research on combating information regarding COVID-19 and predicting news credibility.
In \textbf{CREDBANK}~\cite{mitra2015credbank}, the credibility score is annotated to events, not to news media, for future work, such as the analysis of temporal dynamics of credibility, and social and structural dynamics of events across credibility.
It contains 60 million posts on Twitter covering 96 days, grouped into 1,049 events with 30 annotators' scores on a five-point credibility scale.
Adal{\i} et al. built a multi-labeled dataset composed of news published during the year and the publication media information every year from 2018, called \textbf{NELA-GT}~\cite{nela2018,nela2019,nela2020}.
NELA-GT-2020, for example, includes 1,779,127 articles from 519 news sources, which are categorized in multiple perspectives of veracity as source-level ground truth labels from the Media Bias/Fact Check (MBFC) site~\cite{mbfc}.

Spreading fake news during the 2016 US election shows a strong relationship between fake news and hyperpartisan news~\cite{buzzfeed}.
The researcher and media frequently discuss the possibility of fake news in swaying elections, the impact of fake news on society, and further actions to reduce the impact.
This situation creates the need for datasets that consider news articles with political bias, including hyperpartisan news related to fake news.
\textbf{BFN\_dataset}~\cite{buzzfeed} includes 2,282 posts on Facebook, consisting of 1,145 posts from mainstream pages, 666 from hyperpartisan right-wing pages, and 471 from hyperpartisan left-wing, extracted from nine political news web pages.
Buzzfeed reveals a high affinity between fake news and hyperpartisan pages.
\textbf{MediaReliability}~\cite{baly_2018} includes 1,066 news media information such as a sample of articles, its Wikipedia page, its Twitter account, and so on, for a study on predicting the factuality and bias of news media.
The dataset is also expected to be utilized as prior filtering for fact-checking systems.
\textbf{Buzzfeed-Webis}~\cite{stylometric} is a corpus of 1,627 articles from nine political publishers, three each from the mainstream, the hyperpartisan left, and the hyperpartisan right that have been fact-checked by Buzzfeed.
The dataset is leveraged to model the difference between mainstream and hyperpartisan news, and the detection of hyperpartisan news and fake news.
\textbf{QProp}~\cite{barron2019proppy} is a propaganda dataset composed of 51,294 articles from 104 news media, which are labeled propagandistic (hyperpartisan) or non-propagandistic (mainstream) following Media Bias/Fact Check, for research to detect propaganda news media and to analyze propaganda news articles.

\subsection{Analysis of fake news}
It is necessary to improve the capacity of fake news detection and fact verification but we also need to understand the characteristics of fake news, such as its diffusion into social media ecosystem and its effect on society in order to combat them.
Understanding these causes helps to take efficient measures that reduce the spread of fake news and detect fake news.
Then, the datasets suitable to the research purposes such as  the analysis and understanding of fake news, also are constructed.

Higgs and FibVID were constructed, especially, for the diffusion analysis of fake news on Twitter. 
\textbf{Higgs}~\cite{higgs} a follow-follower network dataset consists of messages posted on Twitter that target during and after the announcement of the discovery of a new particle with the features of the elusive Higgs boson in 2012.
The dataset provides a possible explanation for the observed time-varying dynamics of user activities regarding scientific rumors.
\textbf{FibVID}~\cite{jisu_kim_2021_4441377} is also a network dataset consisting of 1,774 claims labeled in four direction combination of whether a claim is related to COVID-19 and whether a claim is true or fake.
The dataset uncovers the propagation patterns of fake news, specifically during COVID-19 pandemic, and helps to identify the traits of users who engage in fake news diffusion.

\textbf{TSHP-17\_news}~\cite{varying}, includes 14,985 articles from three satire news media, 12,047 articles
from two hoax news media and 33,449 articles from two propaganda news media, and were constructed for the linguistic analysis of each news type, such as satire, hoax, or propaganda.
\textbf{USElection}~\cite{allcott2017social} is composed of 156 fake news stories circulated before the election and extracted from fact-checking sites such as Snopes, for analyzing the 2016 US election.
\textbf{Quote}~\cite{jang2019fake} focuses quote retweet on Twitter, a function of Twitter, to perform an advanced analysis of fake news, while most fake news analysis uses re-share (retweet) information and follow-follower networks.
The COVID-19 epidemic, a global event, provides a strong motivation to build datasets for analyzing fake news related to such events.
For example, \textbf{Misinformation\_COVID-19}~\cite{covid_fakenews} consists of 1,500 posts on Twitter, relating to 1,274 false and 226 partially false claims, comparing the diffusion speed between false and partly false claims.
\textbf{COVID-19\_infodemic}~\cite{infodemic} consists of 586 true news articles and 578 fake news articles to investigate sentiment, article length, and so on.
Memon et al. offer a misinformation dataset, called \textbf{CMU-MisCov19}~\cite{memon2020characterizing}, with 4573 posts on Twitter categorized into 17 themes around the COVID-19 discourse.
\textbf{AraCOVID19-MFH}~\cite{ameur2021aracovid19} is an Arabic COVID-19 dataset with multi-label, designed to consider aspects relevant to the fact-checking task, such as check worthiness, positive/negative label, and factuality, for fake news and hate speech research.

\subsection{Social media posts with images or videos}
The development of our understanding of the fake news ecosystem through analysis and research raises a novel demand for combatting fake news; new tasks are being proposed for real-world applications, and datasets suitable to them are constructed.
One such example is the veracity estimation for social media comments accompanied by doubtful images or videos.
For example, \textbf{PS-Battles}~\cite{ps_battles} was constructed to combat image-manipulation detection.
The dataset includes 102,028 images extracted from Reddit threads, consisting of the original image and a varying number of manipulated derivatives.
\textbf{Fake\_video}~\cite{papadopoulou2019corpus} includes 380 user-generated videos (200 debunked and 180 verified videos) with 5,195 posts and comments on Youtube, Facebook, and Twitter, which are expected to be used to build automated video verification in the future.

\subsection{Guardian}
Recently, ``fact-checking intervention'' in which social media users cite fact-checking websites and reply to fake news spreaders, is a useful strategy to mitigate the spread of fake news~\cite{zhao2015enquiring}.
It is also helpful for verification during the fact-checking process.
This finding encourages research toward the ``Guardian,'' where a user performs the fact-checking intervention themselves.
\textbf{Snopee}~\cite{hannak2014get} includes 3,969 conversational tweets containing links to fact-checking sites during 2012 and 2013.
Fact-checking intervention is called ``snope'' in the paper, and the dataset is utilized for the analysis in terms of who did the act of snoping, friend, follower, the person followed, or stranger in Twitter.
\textbf{Rise-Guardian}~\cite{vo2018rise}, a large-scale version of Snopee, includes 225,068 fact-checking tweets consisting of 157,482 reply tweets and 67,585 retweets collected by Hoaxy~\cite{hoaxy}, which were used for building a fact-checking URL recommendation model to encourage the Guardians to engage more in fact-checking activities.
\textbf{Generate-Factcheck}~\cite{vo2019learning}, which is the extension of Rise-Guardian, includes 73,203 fact-checking tweets including fact-checking articles' URLs with true or false labels and 64,110 original tweets, to generate fact-checking posts relevant to original posts.

\subsection{Check-worthy claims}
New information is created and circulated online at an unprecedented rate, most of which is true and does not need to be verified.
However, we will incur high costs when determining the truthfulness of all the claims that are spread on social media using developed systems, such as fact verification and fake news detection. 
Therefore, the first step in the fact-checking procedure is to identify a claim that is to be fact-checked by consulting reliable sources and is important enough to be worthy of verification, in order to reduce the burden on the verification system.
The task is called as check-worthy claims.
\textbf{ClaimBuster}~\cite{arslan2020benchmark}, which is one of dataset for check-worthy factual claims, consists of 22,281 extracted sentences from the US presidential debate transcripts.
These sentences are categorized into three classes: check-worthy factual sentences, unimportant factual sentences, and non-factual sentences.
CheckThat! lab, one of the competitions to fight misinformation and disinformation in social media and political debates, to organizes a check-worthy claims task as the main task every year since 2018.
For example, two types of English corpora were provided in CheckThat! lab in 2020.
First, \textbf{CT20-English-task1}~\cite{barron2020overview} consisting of 962 posts related to COVID-19 on Twitter.
The other is \textbf{CT20-English-task5}~\cite{barron2020overview}, consisting of 64,290 sentences in 70 transcripts from a political event, debate, or speech.
These datasets are labeled with a binary classification of whether each social media post or political speech is check-worthy.
CT19-T2 and CT20-Arabic introduced in Section 4 that are parts of CheckThat! lab offering datasets are also constructed with the intention of being utilized in check-worthy claims.

\subsection{Claim matching}
The confirmation of whether existing fact-checking sites have already verified a given claim is important to save manual fact-checking efforts and to avoid wasting time.
The confirmation process is necessary because viral claims often come back after a while on social media, and politicians like to repeat their favorite statements, whether true or false.
It is called as ``claim matching.''
Jiang et al. treated the claim matching task as a task of ranking verified claims, which are close to a given claim, and then provided two kinds of datasets: \textbf{KnownLie\_politifact} consisting of 16,636 verified claims from Politifact and \textbf{KnownLie\_snopes} consisting of 10,396 verified claims from Snopes~\cite{shaar2020known}.
Vo et al. collected the dataset of 13,239 positive pairs made by 13,091 posts and 2,170 fact-checking articles for searching fact-checking articles, which address the content of a post containing misinformation, called \textbf{WhereFacts}~\cite{vo2020facts}.
\textbf{WhatsApp-check}~\cite{kazemi2021claim}, which includes pairs made by WhatsApp messages and fact-checked claims, is constructed to explore the effective method to search the suitable claim that matches not only in high-resource languages (English, Hindi), but also in low-resource languages (Bengali, Malayalam, Tamil.)

\subsection{Other datasets}
As for other tasks, for example, \textbf{(Dis)Belief}~\cite{belief} consists of 6,809 posts on Twitter labeled user's disbelief or belief in misinformation for measuring and modeling how much belief each user has by leveraging social media posts.
\textbf{TheTrust-Project}~\cite{zhang2018structured} includes 40 articles of varying credibility annotated with proposed indicators, which is a novel framework for evaluating the credibility of articles.
In addition, several datasets of fake news, in the construction of which authors do not have specific purpose of use, were provided: (e.g., Kaggle\_AN~\cite{kaggle_AN}, Kaggle\_BS~\cite{kaggle_mrisdal}, Hoaxy\_dataset~\cite{hoaxy_data}, FNCorpus~\cite{fnc_}, COVID-19\_dataset~\cite{COVID_fa}, COVID-19\_brazil~\cite{covid-19-brazil})
Research using these datasets is expected in the future.

\begin{table*}[t]
    \centering
    \scriptsize
    \caption{Summary of other datasets related to fake news before 2019.}
    \begin{tabular}{|l|l|c|l|l|l|l|l|l|}
       \hline
       \multirow{2}{*}{\textbf{Dataset}} & \multirow{2}{*}{\textbf{Instances}} & \multirow{2}{*}{\textbf{Labels}} & \multicolumn{1}{c|}{\multirow{2}{*}{\textbf{Topic domain}}} & \multicolumn{1}{c|}{\multirow{2}{*}{\textbf{Raters}}} & \multicolumn{1}{c|}{\textbf{Platforms/}} & \multicolumn{1}{c|}{\multirow{2}{*}{\textbf{Purpose}}} & \multirow{2}{*}{\textbf{Language}} & \multirow{2}{*}{\textbf{Year}}\\ 
        & & & & & \multicolumn{1}{c|}{\textbf{Original data}} & & & \\ \hline
        
        \multirow{2}{*}{Satire~\cite{acl2009_satire}} & \multirow{2}{*}{4,253 articles} & \multirow{2}{*}{2} & \multirow{2}{*}{-} & \multirow{2}{*}{Expert annotators} & English & \multirow{2}{*}{Satire detection} & \multirow{2}{*}{English} & \multirow{2}{*}{2009}\\
        & & & & & Gigaword & & & \\ \hline
        \multirow{2}{*}{Higgs~\cite{higgs}} &  \multirow{2}{*}{985,590 posts} &  \multirow{2}{*}{-} &  \multirow{2}{*}{Higgs boson} &  \multirow{2}{*}{-} &  \multirow{2}{*}{Twitter} &  \multirow{2}{*}{Propagation analysis} & \multirow{2}{*}{English} &  \multirow{2}{*}{2013}\\
        & & & & & & & & \\ \hline
        \multirow{2}{*}{Snopee~\cite{hannak2014get}} & \multirow{2}{*}{3,969 posts} & \multirow{2}{*}{-} & \multirow{2}{*}{Guardian posts} & Fact-checking sites & \multirow{2}{*}{Twitter} & \multirow{2}{*}{Guardian analysis} & \multirow{2}{*}{English} & \multirow{2}{*}{2014}\\
        & & & & (Snopes, Politifact, Factcheck) & & & & \\ \hline
        \multirow{2}{*}{CREDBANK~\cite{mitra2015credbank}} & \multirow{2}{*}{1,379 events} & \multirow{2}{*}{5} & \multirow{2}{*}{-} & \multirow{2}{*}{Expert annotators} & \multirow{2}{*}{Twitter} & \multirow{2}{*}{Event credibility} & \multirow{2}{*}{English} & \multirow{2}{*}{2015}\\
        & & & & & & & & \\ \hline
        \multirow{2}{*}{NewsTrust\_dataset~\cite{cikm2015}} & \multirow{2}{*}{62,064 articles} & \multirow{2}{*}{-} & \multirow{2}{*}{-} & \multirow{2}{*}{NewsTrust} & \multirow{2}{*}{News articles} & \multirow{2}{*}{Media credibility} & \multirow{2}{*}{English} & \multirow{2}{*}{2015}\\
        & & & & & & & & \\ \hline
        \multirow{2}{*}{BFN\_dataset~\cite{buzzfeed}} & \multirow{2}{*}{2,282 posts} & \multirow{2}{*}{4} & \multirow{2}{*}{the 2016 US election} & \multirow{2}{*}{Buzzfeed} & \multirow{2}{*}{Facebook} & \multirow{2}{*}{Analysis} & \multirow{2}{*}{English} & \multirow{2}{*}{2016}\\
        & & & & & & & & \\ \hline
        \multirow{2}{*}{SatiricalCues~\cite{rubin2016fake}} & \multirow{2}{*}{360 articles} & \multirow{2}{*}{2} & Civics, Science, & The Onion, & \multirow{2}{*}{News articles} & \multirow{2}{*}{Satire detection} & \multirow{2}{*}{English} & \multirow{2}{*}{2016}\\
        & & & Business, Soft news & The Beaverton & & & & \\ \hline
        \multirow{2}{*}{TSHP-17\_news~\cite{varying}} & \multirow{2}{*}{60,481 articles} & \multirow{2}{*}{4} & \multirow{2}{*}{-} & \multirow{2}{*}{US News \& World Report} & \multirow{2}{*}{News articles} & \multirow{2}{*}{Linguistic analysis} & \multirow{2}{*}{English} & \multirow{2}{*}{2017}\\
        & & & & & & & & \\ \hline
        \multirow{2}{*}{USElection~\cite{allcott2017social}} & \multirow{2}{*}{156 news} & \multirow{2}{*}{-} & \multirow{2}{*}{the 2016 US election} & Fact-checking sites & \multirow{2}{*}{News articles} & \multirow{2}{*}{Analysis} & \multirow{2}{*}{English} & \multirow{2}{*}{2017}\\
        & & & & (Snopes, Politifact, Buzzfeed) & & & & \\ \hline
        \multirow{2}{*}{Kaggle\_AN~\cite{kaggle_AN}} & \multirow{2}{*}{20,000 posts} & \multirow{2}{*}{3--5} & \multirow{2}{*}{-} & Fact-checking sites & \multirow{2}{*}{Twitter} & \multirow{2}{*}{-} & \multirow{2}{*}{English} & \multirow{2}{*}{2017}\\
        & & & & (Snopes, Politifact, Emergent) & & & & \\ \hline
        \multirow{2}{*}{Kaggle\_BS~\cite{kaggle_mrisdal}} & \multirow{2}{*}{12,999 articles} & \multirow{2}{*}{-} & \multirow{2}{*}{-} & \multirow{2}{*}{BS Detector} & \multirow{2}{*}{News articles} & \multirow{2}{*}{-} & \multirow{2}{*}{English} & \multirow{2}{*}{2017}\\
        & & & & & & & & \\ \hline
        \multirow{2}{*}{Fake.vs.Satire~\cite{fake_satire}} & \multirow{2}{*}{552 claims} & \multirow{2}{*}{2} & \multirow{2}{*}{the US political} & \multirow{2}{*}{Expert annotators} & \multirow{2}{*}{News articles} & \multirow{2}{*}{Satire detection} & \multirow{2}{*}{English} & \multirow{2}{*}{2018}\\
        & & & & & & & & \\ \hline
        \multirow{2}{*}{MediaReliability~\cite{baly_2018}} & \multirow{2}{*}{1,066 media} & \multirow{2}{*}{3} & \multirow{2}{*}{-} & \multirow{2}{*}{MBFC website~\cite{media_bias}} & Wikipedia Pages, & Media bias, & \multirow{2}{*}{English} & \multirow{2}{*}{2018}\\
        & & & & & News articles & Factuality prediction & & \\ \hline
        \multirow{2}{*}{Buzzfeed-Webis~\cite{stylometric}} & \multirow{2}{*}{1,627 articles} & \multirow{2}{*}{3} & \multirow{2}{*}{the 2016 US election} & \multirow{2}{*}{BuzzFeed} & \multirow{2}{*}{News articles} & \multirow{2}{*}{Media style prediction} & \multirow{2}{*}{English} & \multirow{2}{*}{2018}\\
        & & & & & & & & \\ \hline
        \multirow{2}{*}{PS-Battles~\cite{ps_battles}} & \multirow{2}{*}{102,028 images} & \multirow{2}{*}{2} & \multirow{2}{*}{Images} & \multirow{2}{*}{Photoshopbattles subreddit} & \multirow{2}{*}{Reddit} & Image manipulation & \multirow{2}{*}{English} & \multirow{2}{*}{2018}\\
        & & & & & & detection & & \\ \hline
        \multirow{2}{*}{Rise-Guardian~\cite{vo2018rise}} & \multirow{2}{*}{231,377 posts} & \multirow{2}{*}{-} & \multirow{2}{*}{Guardian posts} & \multirow{2}{*}{Hoaxy~\cite{hoaxy}} & \multirow{2}{*}{Twitter} & Guardian & \multirow{2}{*}{English} & \multirow{2}{*}{2018}\\
        & & & & & & recommendation & & \\ \hline
        \multirow{2}{*}{TheTrust-Project~\cite{zhang2018structured}} & \multirow{2}{*}{40 articles} & \multirow{2}{*}{5} & Climate science, & \multirow{2}{*}{Expert annotators} & \multirow{2}{*}{Article credibility} & \multirow{2}{*}{News articles} & \multirow{2}{*}{English} & \multirow{2}{*}{2018}\\
        & & & Public health & & & & & \\ \hline
        \multirow{2}{*}{Hoaxy\_dataset~\cite{hoaxy_data}} & \multirow{2}{*}{20,987,210 posts} & \multirow{2}{*}{3} & \multirow{2}{*}{-} & \multirow{2}{*}{Hoaxy~\cite{hoaxy}} & \multirow{2}{*}{Twitter} & \multirow{2}{*}{-} & \multirow{2}{*}{English} & \multirow{2}{*}{2018}\\
        & & & & & & & & \\ \hline
        \multirow{2}{*}{FNCorpus~\cite{fnc_}} & \multirow{2}{*}{9,400,000 articles} & \multirow{2}{*}{-} & \multirow{2}{*}{News articles} & \multirow{2}{*}{Opensources.co~\cite{opensource}} & \multirow{2}{*}{News articles} & \multirow{2}{*}{-} & \multirow{2}{*}{English} & \multirow{2}{*}{2018}\\
        & & & & & & & & \\ \hline
        \multirow{2}{*}{QProp~\cite{barron2019proppy}} & \multirow{2}{*}{51,294 articles} & \multirow{2}{*}{2} & \multirow{2}{*}{-} & \multirow{2}{*}{Media Bias/Fact Check} & \multirow{2}{*}{News articles} & \multirow{2}{*}{Propaganda Analysis} & \multirow{2}{*}{English} & \multirow{2}{*}{2019}\\
        & & & & & & & & \\ \hline
        \multirow{2}{*}{Quote~\cite{jang2019fake}} & \multirow{2}{*}{3,472 threads} & \multirow{2}{*}{2} & \multirow{2}{*}{-} & \multirow{2}{*}{Kaggle\_BS\cite{kaggle_mrisdal}} & \multirow{2}{*}{Twitter} & \multirow{2}{*}{Analysis} & \multirow{2}{*}{English} & \multirow{2}{*}{2019}\\
        & & & & & & & & \\ \hline
        \multirow{2}{*}{Fake\_video~\cite{papadopoulou2019corpus}} & \multirow{2}{*}{380 videos} & \multirow{2}{*}{2} & Videos & Fact-checking sites & YouTube, Twitter, & Fake video detection & English, Arabic, & \multirow{2}{*}{2019}\\
        & & & with comments & (Snopes, etc.) & Facebook & & French, German & \\ \hline
        \multirow{2}{*}{NELA-GT-2018~\cite{nela2018}} & \multirow{2}{*}{713,534 articles} & \multirow{2}{*}{2--5} & \multirow{2}{*}{News in 2018} & \multirow{2}{*}{Assessment sites} & \multirow{2}{*}{News articles} & Analysis & \multirow{2}{*}{English} & \multirow{2}{*}{2019}\\
        & & & & & & disinformation producer & & \\ \hline
        \multirow{2}{*}{Generate-Factcheck~\cite{vo2019learning}} & \multirow{2}{*}{73,203 posts} & \multirow{2}{*}{-} & \multirow{2}{*}{Guardian posts} & Hoaxy~\cite{hoaxy}, & \multirow{2}{*}{Twitter} & \multirow{2}{*}{Generate guardian} & \multirow{2}{*}{English} & \multirow{2}{*}{2019}\\
        & & & & Rise-Guardian~\cite{vo2018rise} & & & & \\ \hline
    \end{tabular}
    \label{dataset_2019}
\end{table*}

\begin{table*}[t]
    \centering
    \scriptsize
    \caption{Summary of other datasets related to fake news after 2020.}
    \begin{tabular}{|l|l|c|l|l|l|l|l|l|}
       \hline
       \multirow{2}{*}{\textbf{Dataset}} & \multirow{2}{*}{\textbf{Instances}} & \multirow{2}{*}{\textbf{Labels}} & \multicolumn{1}{c|}{\multirow{2}{*}{\textbf{Topic domain}}} & \multicolumn{1}{c|}{\multirow{2}{*}{\textbf{Raters}}} & \multicolumn{1}{c|}{\textbf{Platforms /}} & \multicolumn{1}{c|}{\multirow{2}{*}{\textbf{Purpose}}} & \multirow{2}{*}{\textbf{Language}} & \multirow{2}{*}{\textbf{Year}}\\ 
        & & & & & \multicolumn{1}{c|}{\textbf{Original data}} & & & \\ \hline
        \multirow{2}{*}{NELA-GT-2019~\cite{nela2019}} & \multirow{2}{*}{1,120,000 articles} & \multirow{2}{*}{2--5} & \multirow{2}{*}{News in 2019} & \multirow{2}{*}{Assessment sites} & \multirow{2}{*}{News articles} & Analysis & \multirow{2}{*}{English} & \multirow{2}{*}{2020}\\
        & & & & & & disinformation producer & & \\ \hline
        \multirow{2}{*}{(Dis)Belief~\cite{belief}} & \multirow{2}{*}{6,809 posts} & \multirow{2}{*}{2} & \multirow{2}{*}{-} & Fact-checking site & \multirow{2}{*}{Twitter} & Modeling comment & \multirow{2}{*}{English} & \multirow{2}{*}{2020}\\
        & & & & (Politifact) & & to misinformation & & \\ \hline
        \multirow{2}{*}{ReCOVery~\cite{zhou2020recovery}} & \multirow{2}{*}{2,029 threads} & \multirow{2}{*}{2} & \multirow{2}{*}{COVID-19} & \multirow{2}{*}{Assessment sites} & \multirow{2}{*}{Twitter} & \multirow{2}{*}{Media credibility} & \multirow{2}{*}{English} & \multirow{2}{*}{2020}\\\
        & & & & & &  & & \\ \hline
        \multirow{2}{*}{COVID-19\_infodemic~\cite{infodemic}} & \multirow{2}{*}{1,164 articles} & \multirow{2}{*}{2} & \multirow{2}{*}{COVID-19} & \multirow{2}{*}{Fact-checking sites} & \multirow{2}{*}{News articles} & \multirow{2}{*}{Analysis} & \multirow{2}{*}{English} & \multirow{2}{*}{2020}\\
        & & & & & & & & \\ \hline
        \multirow{2}{*}{Misinformation\_COVID-19~\cite{covid_fakenews}} & \multirow{2}{*}{1,500 posts} & \multirow{2}{*}{2} & \multirow{2}{*}{COVID-19} & Fact-checking sites & \multirow{2}{*}{Twitter} & \multirow{2}{*}{Analysis} & \multirow{2}{*}{English} & \multirow{2}{*}{2020}\\
        & & & & (Snopes, Poynter) & &  & & \\ \hline
        \multirow{2}{*}{CMU-MisCov19~\cite{memon2020characterizing}} & \multirow{2}{*}{4,573 posts} & \multirow{2}{*}{17} & \multirow{2}{*}{COVID-19} & \multirow{2}{*}{Expert annotators} & \multirow{2}{*}{Twitter} & \multirow{2}{*}{Analysis} & \multirow{2}{*}{English} & \multirow{2}{*}{2020}\\
        & & & & & & & &\\ \hline
        \multirow{2}{*}{ClaimBuster~\cite{arslan2020benchmark}} & \multirow{2}{*}{22,281 sentences} & \multirow{2}{*}{3} & US presidential & \multirow{2}{*}{Expert annotators} & \multirow{2}{*}{Speech} & \multirow{2}{*}{Check-worthy claim} & \multirow{2}{*}{English} & \multirow{2}{*}{2020}\\
        & & & debate & & &  & & \\ \hline
        \multirow{2}{*}{CT20-English-task1~\cite{barron2020overview}} & \multirow{2}{*}{962 posts} & \multirow{2}{*}{2} & \multirow{2}{*}{COVID-19} & \multirow{2}{*}{Expert annotators} & \multirow{2}{*}{Twitter} & \multirow{2}{*}{Check-worthy claim} & \multirow{2}{*}{English} & \multirow{2}{*}{2020}\\
        & & & & & &  & & \\ \hline
        \multirow{2}{*}{CT20-English-task5~\cite{barron2020overview}} & \multirow{2}{*}{64,290 sentences} & \multirow{2}{*}{2} & \multirow{2}{*}{Public debate} & Fact-checking site & \multirow{2}{*}{Speech} & \multirow{2}{*}{Check-worthy claim} & \multirow{2}{*}{English} & \multirow{2}{*}{2020}\\
        & & & & (Politifact) & & & & \\ \hline
        \multirow{2}{*}{KnownLie\_politifact~\cite{shaar2020known}} & \multirow{2}{*}{16,636 claims} & \multirow{2}{*}{-} & \multirow{2}{*}{-} & \multirow{2}{*}{Expert annotators} & \multirow{2}{*}{Politifact} & \multirow{2}{*}{Claim matching} & \multirow{2}{*}{English} & \multirow{2}{*}{2020}\\
        & & & & & & & & \\ \hline
        \multirow{2}{*}{KnownLie\_snopes~\cite{shaar2020known}} & \multirow{2}{*}{10,396 claims} & \multirow{2}{*}{-} & \multirow{2}{*}{-} & \multirow{2}{*}{Expert annotators} & \multirow{2}{*}{Snopes} & \multirow{2}{*}{Claim matching} & \multirow{2}{*}{English} & \multirow{2}{*}{2020}\\
        & & & & & & & & \\ \hline
        \multirow{2}{*}{WhereFacts~\cite{vo2020facts}} & \multirow{2}{*}{13,239 pairs} & \multirow{2}{*}{-} & \multirow{2}{*}{-} & Fact-checking sites & \multirow{2}{*}{Twitter} & \multirow{2}{*}{Claim matching} & \multirow{2}{*}{English} & \multirow{2}{*}{2020}\\
        & & & & (Snopes, Politifact) & & & & \\ \hline
        \multirow{2}{*}{COVID-19\_dataset~\cite{COVID_fa}} & \multirow{2}{*}{10,328 headlines} & \multirow{2}{*}{2} & \multirow{2}{*}{COVID-19} & \multirow{2}{*}{-} & \multirow{2}{*}{News articles} & \multirow{2}{*}{-} & \multirow{2}{*}{English} & \multirow{2}{*}{2020}\\
        & & & & & & & & \\ \hline
        \multirow{2}{*}{COVID-19\_brazil~\cite{covid-19-brazil}} & \multirow{2}{*}{1,299 articles} & \multirow{2}{*}{2} & \multirow{2}{*}{COVID-19} & \multirow{2}{*}{Fact-checking sites} & \multirow{2}{*}{News articles} & \multirow{2}{*}{-} & \multirow{2}{*}{Portuguese} & \multirow{2}{*}{2021}\\
        & & & &  & & & & \\ \hline
        \multirow{2}{*}{AraCOVID19-MFH~\cite{ameur2021aracovid19}} & \multirow{2}{*}{10,828 posts} & \multirow{2}{*}{3} & \multirow{2}{*}{COVID-19} & \multirow{2}{*}{Expert annotators} & \multirow{2}{*}{Twitter} & \multirow{2}{*}{-} & \multirow{2}{*}{Arabic} & \multirow{2}{*}{2021}\\
        & & & &  & & & & \\ \hline
        \multirow{2}{*}{FibVID~\cite{jisu_kim_2021_4441377}} & \multirow{2}{*}{1,774 claims} & \multirow{2}{*}{4} & \multirow{2}{*}{COVID-19} & Fact-checking sites & \multirow{2}{*}{Twitter} & \multirow{2}{*}{Propagation analysis} & \multirow{2}{*}{English} & \multirow{2}{*}{2021}\\
        & & & & (Politifact, Snopes) & & & & \\ \hline
        \multirow{2}{*}{NELA-GT-2020~\cite{nela2020}} & \multirow{2}{*}{1,779,127 articles} & \multirow{2}{*}{2--5} & \multirow{2}{*}{News in 2020} & \multirow{2}{*}{Media Bais/Fact Check~\cite{mbfc}} & \multirow{2}{*}{News articles} & Analysis & \multirow{2}{*}{English} & \multirow{2}{*}{2021}\\
        & & & & & & disinformation producer & & \\ \hline
        \multirow{3}{*}{WhatsApp-check~\cite{kazemi2021claim}} &  \multirow{3}{*}{2,333 pairs} &  \multirow{3}{*}{4} & COVID-19, &  \multirow{3}{*}{Expert annotators} &  \multirow{3}{*}{WhatsApp} &  \multirow{3}{*}{Claim matching} & English, Hindi,&  \multirow{3}{*}{2021}\\
        & & & Politics, & & & & Bengail, Tamil, & \\
        & & & Election & & & & Malayalam & \\ \hline
    \end{tabular}
    \label{dataset_2020}
\end{table*}

\section{Discussion and Future Work}
While we have detailed existing datasets on fake news from three perspectives (fake news detection, fact verification, and other tasks) separately in Sections 3--5, they are closely related to the goal of solving the social problems caused by fake news.
In addition, the variety of these datasets and the tasks they target are improving each year.
This can be attributed to the fact that technological advances have increased the number of problems that can be solved. 
Moreover, recent research on fake news has revealed a problem that needs to be solved.
For example, multiple tasks are performed simultaneously, instead of a single task, to more efficiently solve these tasks and share useful features in a single model; specifically, datasets for CheckThat!~\cite{nakov2018overview,elsayed2019overview,barron2020overview} were constructed for solving multiple tasks, check-worthy claims, and fact-verification tasks in the process of fact-checking.
Based on our findings, the important trends and potential tasks for dataset construction in the future have been highlighted.
They include useful insights into the application of the real world or in facilitating the understanding of fake news.

\subsection{Topic domains}
Many datasets with no specific topic domain include the news, depending on whether fact-checking sites have been verified.
However, social events can create a dataset of topic domains that are strongly related to them.
For example, in 2016 and 2017, datasets with political topics were built based on the 2016 US election (BFN\_dataset~\cite{buzzfeed}, Buzzfeed\_political~\cite{benjamin2017}.)
In particular, the COVID-19 pandemic led to the construction of many health-related fake news datasets because ``Infodemic,” the rapid and far-reaching spread of both accurate and inaccurate information about a disease, became a global problem and caused health hazards.
Technical terms appear in a particular topic domain, and technical knowledge is required to determine whether each news item is fake.
Therefore, it is important to build datasets specific to a particular topic domain, such as SCIFACT~\cite{scifact}, which is a new task for verifying the veracity of scientific claims.

\subsection{Languages}
The linguistic characteristics and diffusion patterns of fake news vary by nation and language.
Analyzing fake news across languages leads to the identification of unique non-variant characteristics that are independent of language.
However, the language in most of the existing fake news datasets is English.
The main reason is that, although there is a growing awareness that fact-checking is an important action across the world, there are still not many fact-checking organizations with a sufficient workforce in countries other than the US, the verification in which is the basis for dataset construction.
Furthermore, the infodemic on COVID-19 encourages the construction of a non-English fake news dataset, such as in Hindi~\cite{kar2020no}, Arabic~\cite{COVID-19-FAKES, alam2021fighting}, Chinese~\cite{yang2021checked}, and other multilingual fake news datasets ~\cite{fakecovid, li2020mm}.
In the future, we believe that it is necessary to build fake news datasets in other languages and in a domain other than COVID-19 for the comparison of  languages or nations.

\subsection{Labels}
Many existing datasets are assigned the binary label, fake or real.
This makes it easier to solve fake news detection as a machine learning task by defining it as a binary classification problem.
However, the two labels regarding the truthfulness of each news may be insufficient for detecting fake news that is only partially correct.
For example, Wardle defines seven distinct types of inappropriate content: misinformation and disinformation in satire or parody, false connection, misleading content, false context, imposter content, manipulated content, or fabricated content~\cite{terms1}.
Politifact~\cite{polsite} also judges articles and verifies them based on six ratings: true, mostly true, half true, mostly false, false, or pants on fire.
The flexible labeling of each news item in the dataset is expected to cause the extension of fake news detection to a multi-label classification or regression problem.

\subsection{Functions and Types of platform}
The features used to detect fake news are highly dependent on platform-specific properties.
For example, the suspension of user accounts spreading fake news by platform restriction has the possibility of causing fake news detection from user information not to work properly.
In addition, we may obtain useful features through the platform update; for example, Quote~\cite{jang2019fake} is constructed to examine the quote function toward the spread of fake news introduced by the Twitter update.
In addition, almost all datasets for detecting fake news from social media are constructed based on Twitter, which is a platform for spreading fake news, owing to the convenience of Twitter API.
It is important to build fake news datasets on platforms other than Twitter to understand the differences in the spread of fake news between various platforms, such as Facebook, Reddit, and YouTube.

\subsection{Intention of fake news}
There is the intuition that news created with a malicious intent aims to be more persuasive than those without such aims, and malicious users often play a part in the propagation of fake news to enhance its social influence~\cite{leibenstein1950bandwagon}.
The intuition and definition of fake news introduced in Section 1.1 suggest that it is important to consider the intentions of fake news spreaders.
However, most datasets only follow fact-checking assessments and are not labeled with the consideration of whether there is an intention or not.
In addition to intentions, the degree of harm to society is necessary to determine the priority of fact-checking~\cite{alam2021fighting}.
The annotation of this information, based on features such as intentions, and not only those of whether the news is false or not in the dataset construction, is important for building a highly explainable detection model.

\subsection{Benchmarks of fake news detection}
A benchmark dataset in a certain field can promote the development of technology in the field because it makes a legitimate comparison of the performance between the proposed model and other models: CoNLL 2003 (English)~\cite{con2003} in the field of named entity recognition, 20 Newsgroups~\cite{newsg} in the field of topic models, and COCO Captions~\cite{chen2015microsoft} in the field of image captioning.
In the fact-verification task, the dataset provided by the FEVER workshop\cite{fever1} is treated as a benchmark dataset. 
Twitter15~\cite{twitter15}, Twitter16~\cite{twitter16}, and FakeNewsNet~\cite{shu2020fakenewsnet} in fake news detection tasks are frequently used for the comparison between models, but these are difficult to use as benchmark datasets for the following reasons.
According to the terms and conditions, the social media posts and user information are stored in these datasets as IDs assigned by the social media platform, which we retrieve using the platform API, and subsequently obtain sufficient information.
This means that it is difficult to obtain the dataset at the time it was published because users delete their accounts and posts with the passage of time.
It is incorrect to treat the dataset as a benchmark, the main component of which is social media content, which changes depending on the day of acquisition.
The unchanged dataset construction for fake news detection from social media is an important step toward the development of a benchmark dataset to efficiently verify model performance, with the cooperation of social media platforms.

\subsection{Bias}
An increasing number of studies have been conducted on mitigating bias in a dataset, focusing on the fact that models trained on it can produce biased inference results.
Some papers~\cite{schuster2019towards,suntwal2019importance} investigated whether the inference of the models learned in the FNC\_dataset~\cite{fnc} and FEVER~\cite{fever1}, datasets for a fact-verification task, are biased.
They find the bias of the words in trained models, for example, most of the attention weights by the model are assigned to noun phrases, and propose a mitigation strategy.
In addition to the bias of the words, there may be other biases that have not yet been confirmed in datasets related to fake news, such as author bias~\cite{author_bias}, annotator bias~\cite{annotate_bias}, gender bias~\cite{gender_bias1}, racial bias~\cite{racial_baias1}, and political bias~\cite{poliotics}.
It is important to consider the existence of these biases and their mitigation.

\subsection{Velocity}
Most datasets for fake news detection consist of factual and fake news that actually diffuse over the Internet during a specific period.
The topics and content of datasets built at different times are different because they are strongly influenced by the interests of the general population~\cite{pnas2017}.
This leads to problems in the robustness of the model.
Fake news detection models learned from these datasets achieve high accuracy for the datasets constructed for the same period and domain, while they lead to a drop in the detection performance of fake news in different domains and future applications because of the difference in word appearance. 
For example, a model learned from a dataset in 2017 is difficult to correctly classify articles including ``Donald Trump'' or ``Joe Biden'' in 2021 because the model is not aware of the change in presidents.
As a solution to this problem, a method of replacing the proper nouns in the dataset with information from Wikidata is proposed to make the model more robust~\cite{murayama2021mitigation}.
Also, NELA-GT~\cite{nela2018,nela2019,nela2020} updates its contents every year.
Another possible approach is to propose a detection model with a dynamic knowledge graph.

\section{Conclusion}
Our survey provides extensive reviews of fake news datasets by: (1) summarizing the definition of fake news, relevant concepts related to fake news, and the areas covered by existing survey papers on fake news research as a basis for discussion; (2) introducing 118 datasets related to fake news research on a large scale from three perspectives: 51 datasets for fake news detection, 25 datasets for fact verification, 42 datasets for other tasks, and (3) highlighting challenges in current research and some research opportunities that go with the challenges of the fake news dataset construction.
In fake news research, how and what target dataset to select in the research is an important step to validate the analysis results or the effectiveness of the proposed model.
Our survey facilitates fake news research by helping researchers find suitable datasets without reinventing the wheel and improving fake news studies in depth.


\bibliographystyle{ACM-Reference-Format}
\bibliography{ref}

\end{document}